\documentclass[letterpaper]{article} 
\usepackage{aaai24}  
\usepackage{times}  
\pdfoutput=1
\usepackage{helvet}  
\usepackage{courier}  
\usepackage[hyphens]{url}  
\usepackage{graphicx} 
\usepackage{natbib}  
\usepackage{caption} 
\usepackage[linesnumbered,ruled]{algorithm2e}
\usepackage{multirow,booktabs}
\usepackage{colortbl}
\definecolor{Gray}{gray}{0.9}
\usepackage{subfigure}
\usepackage{bm}
\usepackage{nicefrac}
\usepackage{amsmath}
\usepackage{amsfonts}
\DeclareMathOperator*{\mean}{mean}
\DeclareMathOperator*{\PCA}{PCA}
\DeclareMathOperator*{\argmax}{arg\,max}

\newcommand{\heading}[1]{\noindent\textbf{#1}}

\usepackage{makecell}

\usepackage[switch]{lineno}
\usepackage{soul}
\usepackage{color, xcolor}
\usepackage{pifont}

\newcommand{\ie}{\textit{i.e., }}

\newcommand{\eg}{\textit{e.g., }}
\newcommand{\etc}{\textit{etc}}
\definecolor{best}{HTML}{99FF66}
\definecolor{darkgreen}{HTML}{3F7D31}
\definecolor{darkred}{HTML}{BA3132}

\newcommand{\cmark}{{\ding{51}}}
\newcommand{\xmark}{{\ding{55}}}

\usepackage{appendix}
\title{Contributing Dimension Structure of Deep Feature for Coreset Selection}
\author{
    Zhijing Wan\textsuperscript{\rm 1,2},
    Zhixiang Wang\textsuperscript{\rm 3,4},
    Yuran Wang\textsuperscript{\rm 1,2},
    Zheng Wang\textsuperscript{\rm 1,2}\thanks{Corresponding Author},\\
    Hongyuan Zhu\textsuperscript{\rm 5},
    Shin'ichi Satoh\textsuperscript{\rm 4,3}
}
\affiliations{
    \textsuperscript{\rm 1}National Engineering Research Center for Multimedia Software, Institute of Artificial Intelligence,\\School of Computer Science, Wuhan University, China\\
    \textsuperscript{\rm 2}Hubei Key Laboratory of Multimedia and Network Communication Engineering\\
    \textsuperscript{\rm 3}The University of Tokyo, Japan \\
    \textsuperscript{\rm 4}National Institute of Informatics, Japan\\
    \textsuperscript{\rm 5} Institute for Infocomm Research (I2R) \& Centre for Frontier AI Research (CFAR), A*STAR, Singapore

}

\begin{document}

\maketitle

\begin{abstract}
Coreset selection seeks to choose a subset of crucial training samples for efficient learning. It has gained traction in deep learning, particularly with the surge in training dataset sizes. Sample selection hinges on two main aspects: a sample's representation in enhancing performance and the role of sample diversity in averting overfitting. Existing methods typically measure both the representation and diversity of data based on similarity metrics, such as $L$2-norm. They have capably tackled representation via distribution matching guided by the similarities of features, gradients, or other information between data. However, the results of effectively diverse sample selection are mired in sub-optimality. This is because the similarity metrics usually simply aggregate dimension similarities without acknowledging disparities among the dimensions that significantly contribute to the final similarity. As a result, they fall short of adequately capturing diversity. To address this, we propose a feature-based diversity constraint, compelling the chosen subset to exhibit maximum diversity. Our key lies in the introduction of a novel Contributing Dimension Structure (CDS) metric. Different from similarity metrics that measure the overall similarity of high-dimensional features, our CDS metric considers not only the reduction of redundancy in feature dimensions, but also the difference between dimensions that contribute significantly to the final similarity. We reveal that existing methods tend to favor samples with similar CDS, leading to a reduced variety of CDS types within the coreset and subsequently hindering model performance. In response, we enhance the performance of five classical selection methods by integrating the CDS constraint. Our experiments on three datasets demonstrate the general effectiveness of the proposed method in boosting existing methods.
\end{abstract}

\section{Introduction}

\begin{figure}[!t]
\centering
\includegraphics[width=0.9\columnwidth]{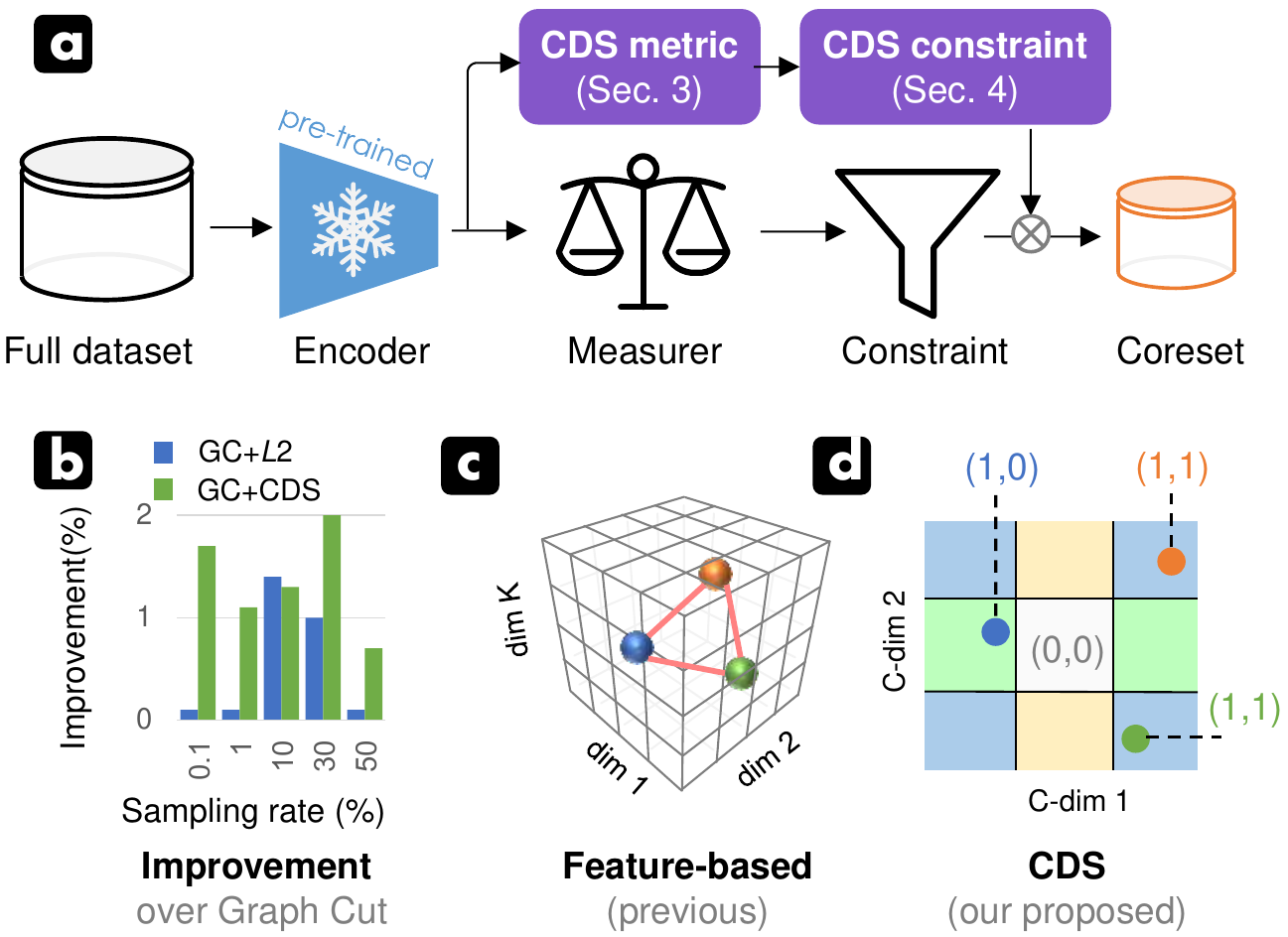}
\caption{Our method and motivation. (a)~We combine the proposed CDS metric and constraint with the current coreset selection pipeline. (b)~CDS metric and constraint enhance the performance of SOTA---GC~\cite{iyer2021submodular}. Although replacing the CDS metric with $L$2 distance employed by previous feature-based methods can improve GC, integrating our proposed CDS metric is more effective since it can capture more diverse, informative samples. (c)~Previous feature-based methods using $L$2 metric could treat three distinct samples as equivalent, while (d)~our CDS metric effectively distinguishes these samples by pruning the feature space and representing the space in different partitions. Note that here, we set the pruned dimension (C-dim) to 2 for demonstration.}
\label{fig:prob}
\end{figure}

Coreset selection is a long-standing learning problem that aims to select a subset of the most informative training samples for data-efficient learning~\cite{das2021tmcoss, wan2023refinement}. Early coreset selection methods were designed to accelerate the learning and clustering of machine algorithms, such as $k$-means and $k$-medians~\cite{har2007smaller}, support vector machines~\cite{tsang2005core}, and Bayesian inference~\cite{campbell2018bayesian}. However, they are designed for specific models and problems, and have limited applications in deep learning. 

Recently, with the rapid development of deep learning~\cite{xie2023striking, yang2023win, yuan2023osap}, research on coreset selection for deep learning has emerged, including geometry-based methods~\cite{sener2017active, agarwal2020contextual}, uncertainty-based methods~\cite{coleman2019selection}, submodularity-based methods~\cite{kothawade2022prism, rangwani2021s3vaada}, gradient matching-based methods~\cite{mirzasoleiman2020coresets, killamsetty2021grad} and others~\cite{toneva2018empirical, swayamdipta2020dataset}. They typically rely on a pre-trained model to obtain information, \eg features, gradients and predicted probabilities, for measuring the importance of data in the full training set through designed measurer and constraint. Budget-sized samples will then be selected based on importance to form a coreset that can be used for data-efficient machine learning~\cite{pooladzandi2022adaptive}, compute-efficient hyper-parameter tuning~\cite{killamsetty2022automata}, continual learning~\cite{tiwari2022gcr, yoon2021online}, \etc. 

The importance of data in coreset selection is assessed in two main aspects, that is, representation and diversity. Representation is guaranteed through the distribution matching between subsets and the full set under the guidance of similarities between data in features, gradients, or other information. Meanwhile, diversity is assured through the imposition of penalties upon similar data. Information similarity is commonly computed by similarity metrics, such as $L$2-norm and cosine distance. However, these similarity metrics obtain the final similarity by simply aggregating dimension similarities without evaluating the impact of different dimensions. This risks treating some distinct samples as equally important, leading to an ineffective assessment of the diversity. For example, in the case of feature-based selection methods, there is redundancy in the feature dimensions extracted by the pre-trained model~\cite{li2017feature}, and similarity metrics directly add a lot of useless information, which is not conducive to measuring the diversity of data. Besides, and most importantly, the data dimensions that contribute significantly to the final similarity are different for different data (such dimensions are called \textit{contributing dimensions}), and this difference is also ignored in the similarity metrics. Our empirical study further shows that similarity metrics tend to select a larger number of samples whose features have the same Contributing Dimension Structure (CDS) for coreset selection, resulting in fewer types of CDS in the coreset, which inhibits model performance. Therefore, it is urgent to design a metric that can evaluate the impact of different dimensions of the data information.

In this paper, we draw inspiration from the above observation and consequently design a Contributing Dimension Structure (CDS) metric and a feature-based CDS diversity constraint (abbreviated as CDS constraint) to improve the current coreset selection pipeline, as shown in Figure~\ref{fig:prob}(a), to compel the selected subset to showcase maximum diversity. Firstly, we propose a CDS metric to select helpful feature dimensions and divide the pruned feature space into different partitions, obtaining the CDS of each data. Throughout our paper, CDS is defined as an indication of whether each dimension of the pruned feature contributes significantly to the overall similarity measure. Each dimension is analyzed in turn, with 1 indicating that the dimension contributes and 0 indicating that it does not. By comparing the CDS of different data, we can classify them as having the same CDS or different CDS. Afterward, an effective strategy \textit{CDS constraint} is proposed to enrich the diversity of CDS in subsets.

The main contribution of our work is threefold:
\begin{itemize}
	\item For the first time in coreset selection, we explicitly introduce information on the Contributing Dimension Structure (CDS) via the proposed \textit{CDS metric} to enrich the diversity of CDS in the coreset.
	\item \textit{CDS constraint}, which aims to constrain the selected subset to have as many different CDS as possible, is proposed to improve existing SOTA methods. We propose two implementations of the CDS constraint, namely the Hard CDS Constraint and the Soft CDS Constraint, and apply them to five classical coreset selection methods.
	\item Extensive experiments on three image classification datasets with two data sampling modes (class-balanced sampling and class-imbalanced sampling) show that our method can effectively improve SOTA methods.
\end{itemize}

\section{Problem Formulation}
\label{sec:problem_def}

\paragraph{Coreset Selection}
We focus on the traditional computer vision task of image classification. In an image classification task with $C$ classes, we work with a sizable training set $\mathcal{D} = {(x_{i}, y_{i})}^n_{i=1}$ defined across a joint distribution $\mathcal{X} \times \mathcal{Y}$, where $n$ denotes the quantity of training data, $\mathcal{X}$ pertains the input space and $\mathcal{Y}$ is the label space $\{1, \ldots, C\}$. In scenarios where there's a specified budget $b$, coreset selection aims to select a subset $\mathcal{S} \subset \mathcal{D}$ containing the most informative training samples. This is done with the intention that the model $\theta^\mathcal{S}$ trained on $\mathcal{S}$ can achieve performance comparable to that of the model $\theta^\mathcal{D}$ trained on the full training set $\mathcal{D}$. The size of the subset $|\mathcal{S}|=b$, where $b < n$. It is common to convert the coreset selection problem into the design of a monotonic objective function $T$ and to find the optimal subset
\begin{equation}
\setlength{\belowdisplayskip}{3pt}
\mathcal{S}^{*} = \argmax_{\mathcal{S} \subset \mathcal{D}}\, T\left(\mathcal{S}\right) \quad \text{s.t.} \quad \left|\mathcal{S}\right| \leq b,
\label{eq:S}
\end{equation}
where budget-sized $\mathcal{S}^{*}$ is selected before training, with the expectation that the accuracy of models trained on this subset will be maximized.

Representativeness and diversity of subsets are the two main factors that coreset selection focuses on when measuring the importance of data. Most existing coreset selection methods usually design and implement the objective function based on similarity metrics (\eg $L$2-norm). They ensure representativeness through distribution matching guided by similarities of features, gradients, or other information between data, while pursuing diversity by penalising data with similar information. However, conventional similarity metrics simply aggregate dimension similarities without acknowledging disparities among the dimensions that significantly contribute to the final similarity. As a result, they fall short of adequately capturing diversity. We will dissect the problems in the following.

\begin{figure}
    \centering
    \includegraphics[width=\columnwidth]{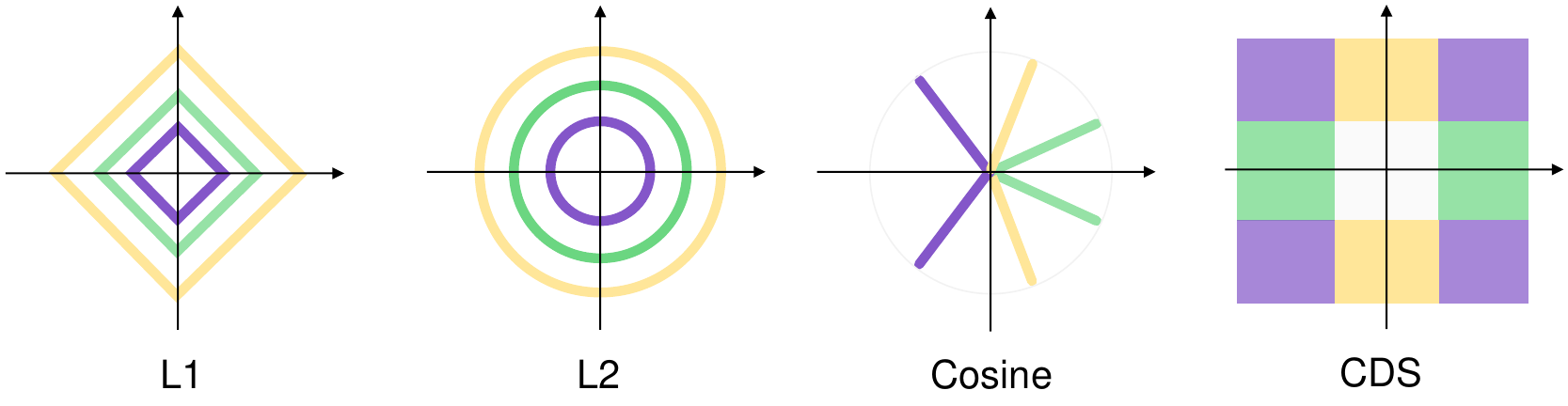}
    \caption{Different metrics. {$L$1 and $L$2 quantify the magnitude between samples, while the Cosine distance evaluates the direction between samples. In contrast, our introduced CDS metric evaluates the impact of different dimensions, making it an effective tool for assessing diversity. Note that the reference sample is positioned at the origin for $L$1, $L$2, and CDS, while it is located along the positive horizontal axis for Cosine distance. Regions with the same color indicate the same score. 
}}
    \label{fig:diversity}
\end{figure}

\paragraph{Diversity Measurement}
Informative and easily accessible deep features are often adopted in selection methods~\cite{guo2022deepcore, margatina2021active, wan2023multi, wan2022efficient}, which typically use the overall similarity metrics to calculate the similarity between deep features to measure the importance of the data, such as $L$1-norm, $L$2-norm and cosine distance metric. Figure~\ref{fig:diversity} illustrates their characteristics. For convenience, we focus on the most commonly adopted $L$2-norm as an example. It can be formulated as:
\begin{equation}
d\left(\bm{F}(x_{i}), \bm{F}(x_{j})\right)=\sqrt{\sum_{k=0}^{K-1}\left(f_{i}^{k}-f_{j}^{k}\right)^2},
\label{eq:l2-norm}
\end{equation}
where $\bm{F}(x_{i})=\left[f_{i}^0, f_{i}^1, \ldots, f_{i}^{K-1}\right] \in \mathbb{R}^{K}$ denotes the feature of sample $x_{i}$ extracted by the model, and $K$ means the number of dimensions of the deep feature. 

Unfortunately, the calculation of the similarity metric shown in Equation~\ref{eq:l2-norm} does not explicitly reflect the numerical difference between $\bm{F}(x_{i})$ and $\bm{F}(x_{j})$ in \emph{each} dimension. This may lead to an ineffective assessment of the diversity of the data, making the selection of the coreset insufficiently diverse. For example, there is a phenomenon: Given candidate data $x_{i}$, $x_{j}$, and $x_{h}$, when measuring their importance, we compare them with a selected data $x_{q}$. If their distance $d\left(\bm{F}(x_{i}), \bm{F}(x_{q})\right)$, $d\left(\bm{F}(x_{j}), \bm{F}(x_{q})\right)$, and $d\left(\bm{F}(x_{h}), \bm{F}(x_{q})\right)$ are close,
feature-based methods could treat them as equally important and select them with equal probability. However, when analyzed in terms of each dimension difference, there may exist a dimension $\ddot{k}$ that makes the difference between $x_i$ and $x_q$ equals zero, while that between $x_j$ and $x_q$, $x_h$ and $x_q$ are significantly larger than zero, \ie $$|f_{i}^{\ddot{k}}-f_{q}^{\ddot{k}}| \to 0\,,\,\,\,\, |f_{j}^{\ddot{k}}-f_{q}^{\ddot{k}}| \gg 0\,,\,\,\,\, |f_{h}^{\ddot{k}}-f_{q}^{\ddot{k}}| \gg 0\,.$$
This can be interpreted as the $\ddot{k}^{th}$ dimension of data $x_{i}$ does not contribute in calculating the overall similarity with $x_{q}$, while the $\ddot{k}^{th}$ dimension of data $x_{j}$ and data $x_{h}$ do. In other words, $x_{i}$ has a different contributing dimension structure (CDS) from $x_{j}$ and $x_{h}$. At this point, the CDS diversity of $x_{i}$ is different from that of $x_{j}$ and $x_{h}$ while their diversities calculated by $L$2 are the same. 

It is natural to ask the \textbf{question}: {what is the relationship between the diversity of the CDSs in the selected samples and the performance of the models trained on those selected samples?}
In light of this question, we introduce a metric to measure the CDS of each data in the next section.

\begin{figure}[t]
\centering
\includegraphics[width=\columnwidth]{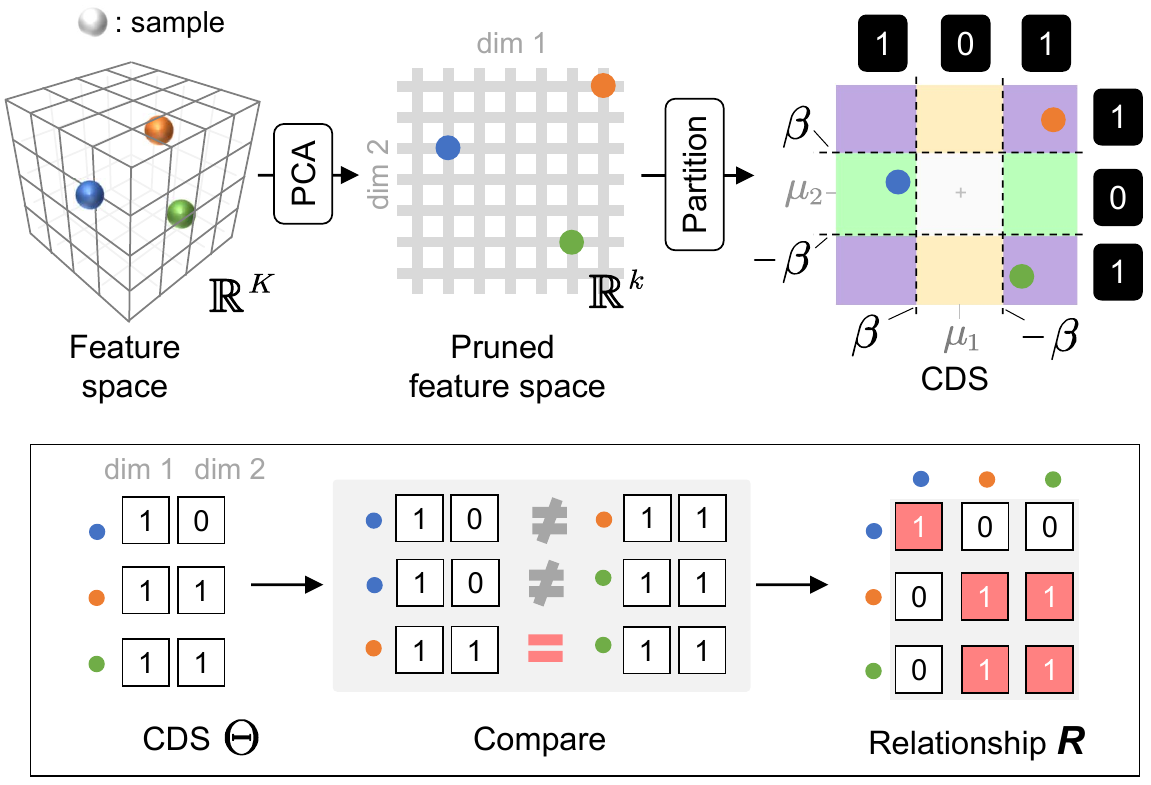}
\caption{CDS metric for deep features. Given the high-dimensional feature matrix, we first reduce its dimension from $K$ to $k$ using PCA. Then, we compute the central feature $[\mu_{0}$, $\mu_{1}$, \dots, $\mu_{k-1}]$ of the dimension reduced feature matrix. Next, we obtain the CDS for each data by comparing the difference between each data feature and the central feature in each dimension with a threshold $\beta$ to divide the feature space into different partitions. Finally, the CDS relationship matrix $\bm{R}$ used for the subsequent CDS constraint is obtained by comparing the CDS between each data individually to see if they are the same.
}
\label{fig:CDS_metric}
\end{figure}

\section{CDS Metric of Deep Feature}
\label{sec:CDS_DF}

The previous section introduces the concept of CDS. Moving forward, we will introduce the CDS metric in this section. This metric is designed to quantify the CDS of individual samples, as depicted in Figure~\ref{fig:CDS_metric}. Additionally, we will delve into an analysis of the connection between CDS diversity and the performance of models.

\subsection{CDS Metric}
\label{sec:CDS}

\paragraph{Dimension Reduction}
Inspired by the work~\cite{li2017feature}, we consider that for the analysis of CDS relationships between data, there is redundancy in the feature dimensions output by the network, \ie the dimensions that are helpful need to be selected before analysis. Therefore, we first perform dimension selection using the classical PCA algorithm~\cite{pearson1901liii} to reduce $K$ to $k$. The choice of $k$ will be elaborated upon in Section~{\ref{ablation}}. After that, two problems lie ahead when it comes to actually analyzing the CDS relationships between the data.

\paragraph{Deviation from the Mean}
One problem is that the relationship between the contributing dimension structures of the data is relative, \ie in the above phenomenon, the contributing dimension structure relationship between $x_{i}$, $x_{j}$ and $x_{h}$ may change as $x_{q}$ changes to $x_{\dot{q}}$. Therefore in sampling data from each class, we set $\bm{F}(x_{q})$ to be the class prototype, \ie the central feature $\tilde{\bm{F}_{c}}=[\mu_{0}, \mu_{1}, \dots, \mu_{k-1}]$ of the class~\cite{xie2022sampling}, to consistently analyze the relationships of all training data\footnote{For class-imbalanced sampling settings, $\bm{F}(x_{q})$ can be set to the central feature of the dataset.} with $$\mu_{k'}=\left(\sum_{i=0}^{N_{c}-1}{f_{i}}^{k'}\right)/N_{c}\,, \,\, \, k' \in \{0, 1, \dots, k-1\}\,.$$ Then for each data point within class $c$, its deviation from the central feature $\tilde{\bm{F}}_{c}$ in each dimension can be expressed as $\bm{\sigma} = \left[|f_{i}^{0}-\mu_{0}|, |f_{i}^{1}-\mu_{1}|, \ldots, |f_{i}^{k-1}-\mu_{k-1}|\right] \in \mathbb{R}^{k}$, where $i \in \{0, 1, \ldots, N_{c}-1 \}$.

\paragraph{Partition}
The other problem is determining if each dimension contributes to the similarity calculation. We intuitively set a threshold parameter $\beta$ to deal with it. Then, the CDS $\Theta$ of data $x_{i}$ can be represented as:
\begin{equation}
\Theta(x_{i}) = \left[\mathbb{I}(|f_{i}^{0}-\mu_{0}|), \ldots, \mathbb{I}(|f_{i}^{k-1}-\mu_{k-1}|)\right],
\label{eq:CDS}
\end{equation}
where $\mathbb{I}(\triangle f)$ is a binary decision function. When the feature difference $\triangle f > \beta$, then $\mathbb{I}(\triangle f) = 1$, otherwise $\mathbb{I}(\triangle f) = 0$. 

\paragraph{Comparison}
After that, the relationship between data can be subdivided into two types from the CDS perspective. Take $x_{i}$ and $x_{j}$ as an example: 
\begin{itemize}
    \item if $\forall k' \in \{0, 1, \ldots, k-1\}$, $\mathbb{I}(|f_{i}^{k'}-\mu_{k'}|) = \mathbb{I}(|f_{j}^{k'}-\mu_{k'}|)$, then $x_{i}$ and $x_{j}$ have the same contributing dimension structure, noted as $R_{ij}=1$;
    \item if $\exists k'\in \{0, 1, \ldots, k-1\}$, $\mathbb{I}(|f_{i}^{k'}-\mu_{k'}|) \neq \mathbb{I}(|f_{j}^{k'}-\mu_{k'}|)$, then $x_{i}$ and $x_{j}$ have different contributing dimension structures, noted as $R_{ij}=0$.
\end{itemize}
To achieve this division, we adopt the cosine similarity to measure the similarity between the CDS of $x_{i}$ and the CDS of $x_{j}$. If the cosine similarity of $\Theta(x_{i})$ and $\Theta(x_{j})$ is equal to 1, then it means that $x_{i}$ and $x_{j}$ have the same CDS; otherwise, it means that $x_{i}$ and $x_{j}$ have different CDS. Following this approach, the CDS relationship between each data is calculated and analyzed individually. Then, we can obtain a relationship matrix $\bm{R}^{c} \in \mathbb{R}^{N_{c} \times N_{c}}$ for the class $c$, where $R^{c}_{ij} \in \{0, 1\}$. The relationship matrix $\bm{R}^{c}$ will be used in the subsequent CDS Constraint algorithm.

\subsection{Analyses}
\label{sec:analyze}

\begin{figure}
    \centering
    \includegraphics[width=\columnwidth]{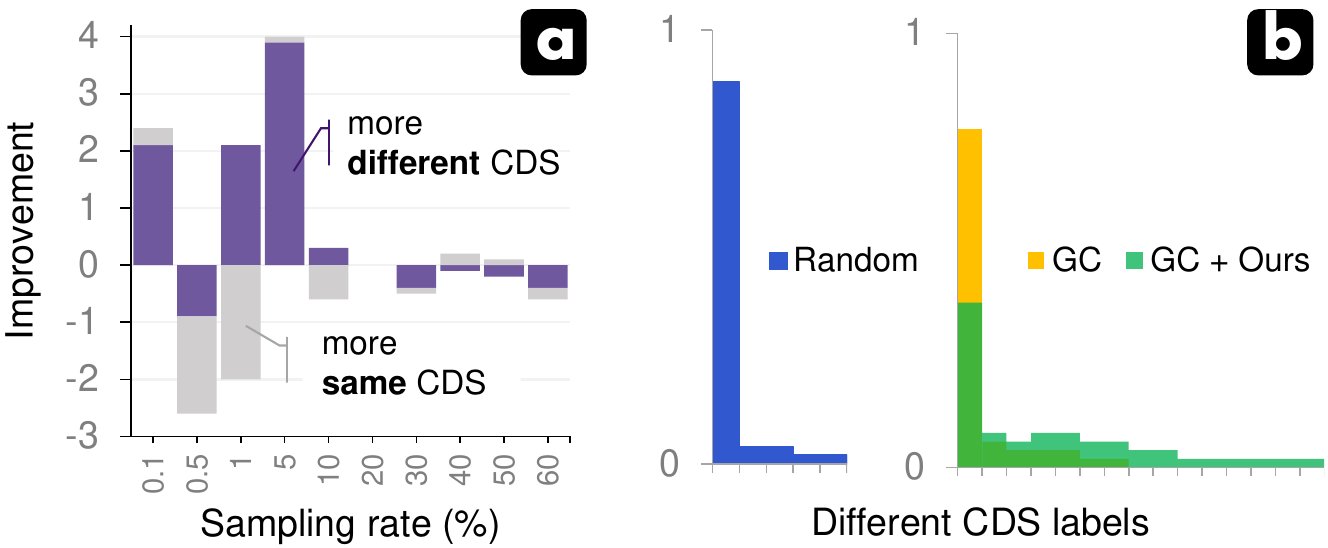}
    \caption{Analyses. (a) The improvement of sampling more same CDS strategy or more different CDS strategy over random sampling. The strategy of sampling more of the same CDS performs worse than random sampling. The strategy of sampling more different CDSs performs better than random sampling, especially with low sampling rates of 0.1\%-10\%. The result motivates us to select more samples with different CDS. (b) We compare the CDS distribution of coresets (1\% of the CIFAR-10) selected by the baseline method and our improved counterparts. It exhibits that previous methods tend to choose a few certain CDSs, which could lead the trained model to perform worse than random sampling. Integrating our proposed constraint explicitly increases the diversity of CDS in the selected coreset.
}
\label{fig:analysis}
\end{figure}

We performed experiments to analyze the connection between CDS diversity and the performance of models. Firstly, we computed the CDS relationship matrix for each class in the dataset using the CDS metric. Then, guided by the relationship matrix, two classes of data were sampled according to the sampling rate, \ie more data with the same CDS (more S-CDS) and more data with different CDS (more D-CDS). They were used to train the models separately and then we compared their performances with that of the Random method. Please refer to the \emph{Supplementary Material \ref{sec:details_moreSD}} for details of the experimental setup. Figure~\ref{fig:analysis}(a) shows the comparison results. It shows that the more D-CDS strategy outperforms the more S-CDS strategy when sampling 0.1\%-10\% of CIFAR-10; the two strategies are evenly matched as the sampling rate increases. Therefore, for data with the same overall similarity, selecting a subset with different CDS impacts performance differently than selecting a subset with the same CDS at low sampling rates. Specifically, more data with different CDSs need to be sampled.

When further using the CDS metric to analyze the CDS relationships of the data sampled by existing SOTA methods, we find that they tended to select data with the same CDS, as evidenced in Figure~\ref{fig:analysis}(b). With reference to as shown in Figure~\ref{fig:analysis}(a), it can be deduced that the coresets selected by the existing SOTA methods are sub-optimal.

\section{Coreset Selection with CDS Constraints}
\label{sec:CS_CDS}
Inspired by the results in Figure~\ref{fig:analysis}, in this section, we propose to improve the SOTA methods by using a feature-based CDS diversity constraint (a.k.a. CDS constraint). The key idea of the CDS constraint is to sample a subset with as many different CDS as possible. We present two implementations of the CDS constraint: the \emph{hard} version, which can be directly applied to existing methods, and the \emph{soft} version, which requires custom design aligned with the objective function of the targeted coreset selection methods.

\subsection{Implementation I: Hard CDS Constraint}
\label{sec:hard}
The coreset selection algorithm with Hard CDS Constraint consists of a two-stage clustering process and a data selection process. The data are first clustered based on the feature distance and CDS relationships. Then, the data are selected using a baseline selection method among data clusters with the same feature distance and CDS. We depict the details in the following:
\begin{enumerate}
    \item \textbf{1$^\text{st}$ stage clustering:} the distance between each reduced feature and the central feature (of the class or dataset) is calculated and the data are clustered according to the spacing $\alpha$. That is, samples with feature distance values $ v \in [h \times \alpha, (h+1)\times \alpha)$ are clustered into one group. In our experiments, $\alpha = 0.5$ and $h \in N^+$. At this point, the sampling budget for each cluster is calculated based on the cluster density. \item \textbf{2$^\text{nd}$ stage clustering:} for each cluster in the 1$^\text{st}$ stage, the CDS relationship between the data within the cluster is calculated and the data with the same CDS are re-clustered to obtain multiple clusters $t$. At this point, to achieve the CDS constraint, we constrain the sampling budget for each cluster $t$ to be as consistent as possible. 
    \item \textbf{data selection}: each cluster $t$ is then sampled according to the baseline method. Please refer to the \emph{Supplementary Material \ref{sec:cs_hard CDS}} for a detailed algorithm of coreset selection with Hard CDS Constraint.
\end{enumerate}

\subsection{Implementation II: Soft CDS Constraint}
\label{sec:soft}
Although Hard CDS Constraint does not need to be redesigned and can be used directly with any of the coreset selection methods, in practice, it has been found to improve only some methods (see the \emph{Supplementary Material \ref{sec_hard_vs_soft}} for related experiments). This means that for other methods, a special design is required. For this reason, we have further proposed the Soft CDS Constraint, which is integrated into the objective function $T$ of existing methods, \ie CRAIG and Graph Cut. We expect to constrain the original objective function $T$ by designing a constraint function $H$ related to the relationship matrix $\bm{R}$ to find:
\begin{equation}
\begin{gathered}
\mathcal{S}^{\star} = \argmax_{\mathcal{S} \subset \mathcal{D}} T\left(\mathcal{S}\right) \times H(\bm{R}) \quad \text {s.t.} \quad \left|\mathcal{S}\right| \leq b,
\label{eq:S_CDS}
\end{gathered}
\end{equation}
where $\mathcal{S}^{\star}$ satisfies the CDS constraint:
\begin{equation}
\Psi(\mathcal{S}^{\star}) \ge \Psi(\mathcal{S}^{*}).
\label{eq:CDS_constraint}
\end{equation}
$\Psi(\mathcal{S})$ calculates the number of CDS types in subset $\mathcal{S}$.

Since finding the optimal subset $\mathcal{S}^{\star}$ is NP-hard in general, we use the simple greedy algorithm~\cite{minoux2005accelerated} to approximately solve the coreset selection problem, as in the CRAIG and Graph Cut. The greedy algorithm starts with the empty set $\mathcal{S}_0 = \emptyset$, and at each selection iteration $i$, it selects an element $e \in V$ that maximizes $T\left(e|\mathcal{S}_{i-1}\right) \times H(\bm{R})$ and enriches the diversity of CDS, \ie $\mathcal{S}_{i} = \mathcal{S}_{i-1} \cup {\argmax_{e \in V}T\left(e|\mathcal{S}_{i-1}\right) \times H(\bm{R})}$, where $V = \mathcal{D} \setminus \mathcal{S}_{i-1}$ and $\Psi(\mathcal{S}_{i}) \ge \Psi(\mathcal{S}_{i-1})$. Due to page limitations, the detailed algorithm is provided in the Algorithm~\ref{alg:algorithm_soft} of \emph{Supplementary Material}.

\paragraph{CRAIG with the CDS Constraint} \
\label{sec:CRAIG_CDS}
CRAIG~\cite{mirzasoleiman2020coresets} is a gradient matching based method that tries to find an optimal coreset $\mathcal{S}$ that approximates the full dataset gradients under a maximum error $\epsilon$ by converting gradient matching problem to the maximization of a monotone submodular facility location function $T$. It uses the greedy algorithm to select data. To satisfy the CDS constraint, we design a function $H$ to constrain the monotone submodular facility location function:
\begin{equation}
e = \argmax_{i \in \mathcal{D} \setminus \mathcal{S}_{l-1}}T\left(i|\mathcal{S}_{l-1}\right) \times H(\bm{R})_{i},
\end{equation}
our designed constraint function $H$ for CRAIG is:
\begin{equation}
H\left(\bm{R}\right)_{i} = {1}/{\left(\sum_{j \in \mathcal{S}_{l-1}}R_{ij}+1\right)}.
\end{equation}

\paragraph{Graph Cut with the CDS Constraint} \
\label{sec:GC_CDS}
Graph Cut~\cite{iyer2021submodular} is a submodularity-based method that naturally measures the information and diversity of the selected subset $\mathcal{S}$. At selection iteration $l$, a greedy algorithm is used to find:
\begin{equation}
\footnotesize
e = \argmax_{i \in \mathcal{D} \setminus \mathcal{S}_{l-1}}\left(\sum_{o \in \mathcal{D}}s(G_{i}, G_{o}) - \lambda \times \sum_{j \in \mathcal{S}_{l-1}}s(G_{i}, G_{j}) \right),
\end{equation}
where $s(\cdot, \cdot)$ is a similarity metric that measures the gradient similarity between data. 
The parameter $\lambda$ captures the trade-off between diversity and representativeness, and $\lambda=2$ in our implementation. Considering the item $\lambda \times \sum_{j \in \mathcal{S}_{l-1}}s(G_{i}, G_{j})$ is responsible for measuring the diversity of each data, we thus propose that the CDS diversity constraint function $H$ constrains only this item: $\lambda \times \sum_{j \in \mathcal{S}_{l-1}}s(G_{i}, G_{j}) \times H(\bm{R})_{ij}$. Our designed constraint function $H$ is given by:
\begin{equation}
H\left(\bm{R}\right)_{ij}=\left\{\begin{array}{c} 2, \qquad \quad \text{if}~~ R_{ij}=1 \\
1, \qquad \quad \text{if}~~ R_{ij}=0
\end{array}\right. .
\end{equation}
If data $x_{i}$ have the same CDS as the data $x_{j}$ from $\mathcal{S}_{l-1}$, $H\left(\bm{R}\right)_{ij}$ can increase the penalty value, reducing the probability of data $x_i$ being selected.

\section{Experiments}

\subsection{Datasets, Model and Experimental Setup}
We evaluate our method with two common data sampling modes, \ie class-balanced and class-imbalanced sampling. We perform experiments on three common image classification datasets, including CIFAR-10, CIFAR-100~\cite{krizhevsky2009learning}, and TinyImageNet (TIN, a subset of ImageNet~\cite{krizhevsky2012imagenet}). We provide details of datasets in the \emph{Supplementary Material \ref{sec:datasets_details}}.

\begin{table*}[!t]
\small
\centering
\begin{tabular}{clcccccc}
    \toprule
        \multicolumn{2}{c}{\multirow{2}{*}{\textbf{Method}}} & \multicolumn{6}{c}{\textbf{Sampling rates}}\\ \cmidrule(lr){3-8}
        & & 0.1\% & 0.5\% & 1\% & 5\% & 10\% & 20\% \\ \midrule
        \multicolumn{2}{c}{Random} & 18.9{$\pm$0.2} & 29.5{$\pm$0.4} & 39.3{$\pm$1.5} & 62.4{$\pm$1.7} & 74.7{$\pm$1.9} & 
        {86.9}{$\pm$0.3} \\ 
        \multicolumn{2}{c}{KCG} & 18.7{$\pm$2.9} & 27.4{$\pm$1.0} & 31.6{$\pm$2.1} & 53.5{$\pm$2.9} & 73.2{$\pm$1.3} & 86.9{$\pm$0.4} \\

        \multicolumn{2}{c}{Forgetting} & 21.8{$\pm$1.7} & 29.2{$\pm$0.7} & 35.0{$\pm$1.1} & 50.7{$\pm$1.7} & 66.8{$\pm$2.5} & 86.0{$\pm$1.2} \\
    
        \multicolumn{2}{c}{LC} & 14.8{$\pm$2.4} & 19.6{$\pm$0.8} & 20.9{$\pm$0.4} & 37.4{$\pm$1.9} & 56.0{$\pm$2.0} & 83.4{$\pm$1.1} \\

        \multicolumn{2}{c}{CRAIG} & 21.1{$\pm$2.4} & 27.2{$\pm$1.0} & 31.5{$\pm$1.5} & 45.0{$\pm$2.9} & 58.9{$\pm$3.6} & 79.7{$\pm$3.5} \\

        \multicolumn{2}{c}{Cal} & 20.8{$\pm$2.8} & 32.0{$\pm$1.9} & 39.1{$\pm$3.2} & 60.7{$\pm$0.8} & 72.2{$\pm$1.5} & 79.9{$\pm$0.5} \\

        \multicolumn{2}{c}{Glister} & 19.5{$\pm$2.1} & 29.7{$\pm$1.1} & 33.2{$\pm$1.1} & 47.1{$\pm$2.6} & 65.7{$\pm$1.7} & 83.4{$\pm$1.7} \\

        \multicolumn{2}{c}{GC} & 22.9{$\pm$1.4} & 34.0{$\pm$1.3} & 42.0{$\pm$3.0} & 66.2{$\pm$1.0} & 75.6{$\pm$1.4} & 84.3{$\pm$0.4} \\

        \multicolumn{2}{c}{M-DS} & 21.0{$\pm$3.0} & 31.8{$\pm$1.2} & 37.7{$\pm$1.4} & 63.4{$\pm$2.2} & 78.0{$\pm$1.3} & 87.9{$\pm$0.5} \\ \midrule
        
        \multicolumn{2}{c}{GC+Ours} & {\textbf{24.6}}{$\pm$\textbf{1.7}}  & {\textbf{36.4}}{$\pm$\textbf{1.0}}  & {\textbf{43.1}}{$\pm$\textbf{1.8}}  & {\textbf{67.1}}{$\pm$\textbf{0.6}}  & {76.9}{$\pm$0.2}  & 85.2{$\pm$0.6}  \\
        
        \multicolumn{2}{c}{$\Delta$} & {1.7 $\uparrow$} & {2.4 $\uparrow$} & {1.1 $\uparrow$} & {0.9 $\uparrow$} & {1.3 $\uparrow$} & {0.9 $\uparrow$} \\ \midrule

        \multicolumn{2}{c}{M-DS+Ours} & {22.0}{$\pm$2.0}  & {33.0}{$\pm$1.3}  & {40.7}{$\pm$1.0}  & {64.9}{$\pm$0.8}  & {\textbf{79.6}}{$\pm$\textbf{0.4}}  & {\textbf{87.9}}{$\pm$\textbf{0.2}}  \\
        
        \multicolumn{2}{c}{$\Delta$} & {1.0 $\uparrow$} & {1.2 $\uparrow$} & {3.0 $\uparrow$} & {1.5 $\uparrow$} & {1.6 $\uparrow$} & {0.0 $\uparrow$} \\
        
        \bottomrule
    \end{tabular}
\captionof{table}{Comparison on the class-balanced sampling setting. We train randomly initialized ResNet-18 on coresets of CIFAR-10 selected by different methods and then test them on the test set of CIFAR-10. Bold emphasizes the best performance at each sampling rate. $\Delta$ denotes the improvement of baseline+Ours over baseline.}
\label{tab:CDS_CIFAR10}
\end{table*}

In all experiments, we utilize the 18-layer residual network (ResNet-18)~\cite{he2016deep} as the backbone of the pre-trained model and target model, and use the deep features extracted before the final fully connected layer for the CDS metric, which is 512-dimensional (512-D). We follow the experimental setup of work~\cite{guo2022deepcore}. Specifically, we use SGD as the optimizer with batch size 128, initial learning rate 0.1, Cosine decay scheduler, momentum 0.9, weight decay $5 \times 10^{-4}$, 10 pre-trained epochs and 200 training epochs. For data augmentation, we apply random crop with 4-pixel padding and random flipping on the $32\times32$ training images. We use classification accuracy as the evaluation metric. For each selection method, we repeat the same experiment 5 times with random seeds and report the performance mean and standard deviation. All experiments were run on Nvidia Tesla V100 GPUs.

\subsection{Comparison Methods}
We reproduce nine selection methods ourselves based on the open source database\footnote{https://github.com/PatrickZH/DeepCore}, including Random, K-Center Greedy (KCG)~\cite{sener2017active}, Forgetting~\cite{toneva2018empirical}, Least Confidence (LC)~\cite{coleman2019selection}, CRAIG~\cite{mirzasoleiman2020coresets}, Cal~\cite{margatina2021active}, Glister~\cite{killamsetty2021glister}, Graph Cut (GC)~\cite{iyer2021submodular}, and Moderate-DS (M-DS)~\cite{xia2023moderate}. To adequately demonstrate the effectiveness of the CDS constraint, we apply the CDS constraint to four classical coreset selection methods of different types and an up-to-date selection method (each as a baseline): 
\begin{itemize}
    \item \textbf{KCG ---} a feature distribution-based selection method; 
    \item \textbf{LC ---} an uncertainty-based selection method; 
    \item \textbf{CRAIG ---} a gradient matching-based selection method;
    \item \textbf{GC --- } a submodularity-based selection method;
    \item \textbf{M-DS ---} a feature distribution-based selection method.
\end{itemize}
We improve baseline methods KCG, LC, and M-DS using the Hard CDS constraint and the greedy sampling baselines CRAIG and GC using the Soft CDS constraint. 

\begin{figure}[!t]
\centering
\subfigure[KCG]{
\label{fig:f_a}
\includegraphics[width=0.305\columnwidth]{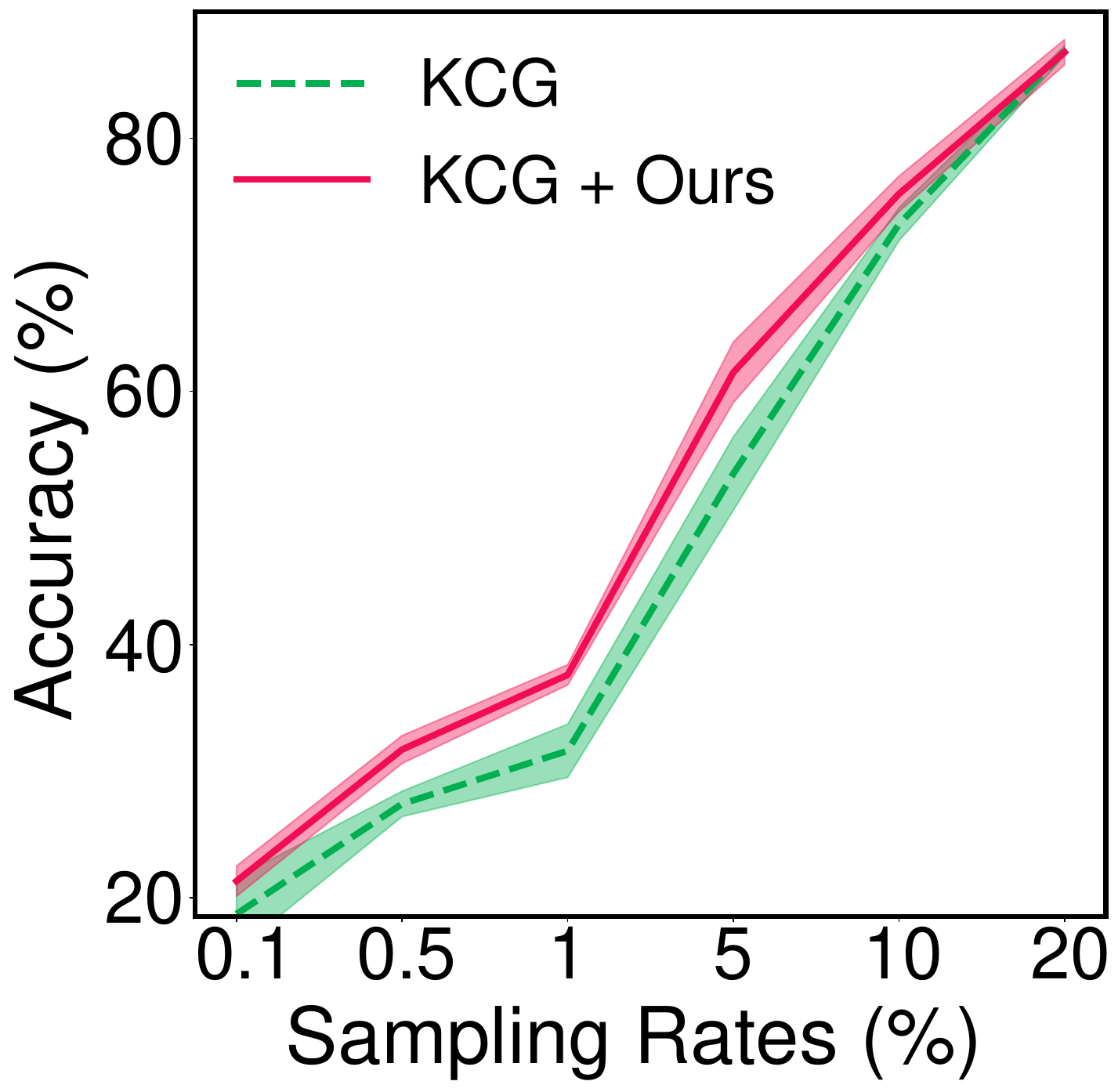}}
\subfigure[LC]{
\label{fig:f_b} 
\includegraphics[width=0.305\columnwidth]{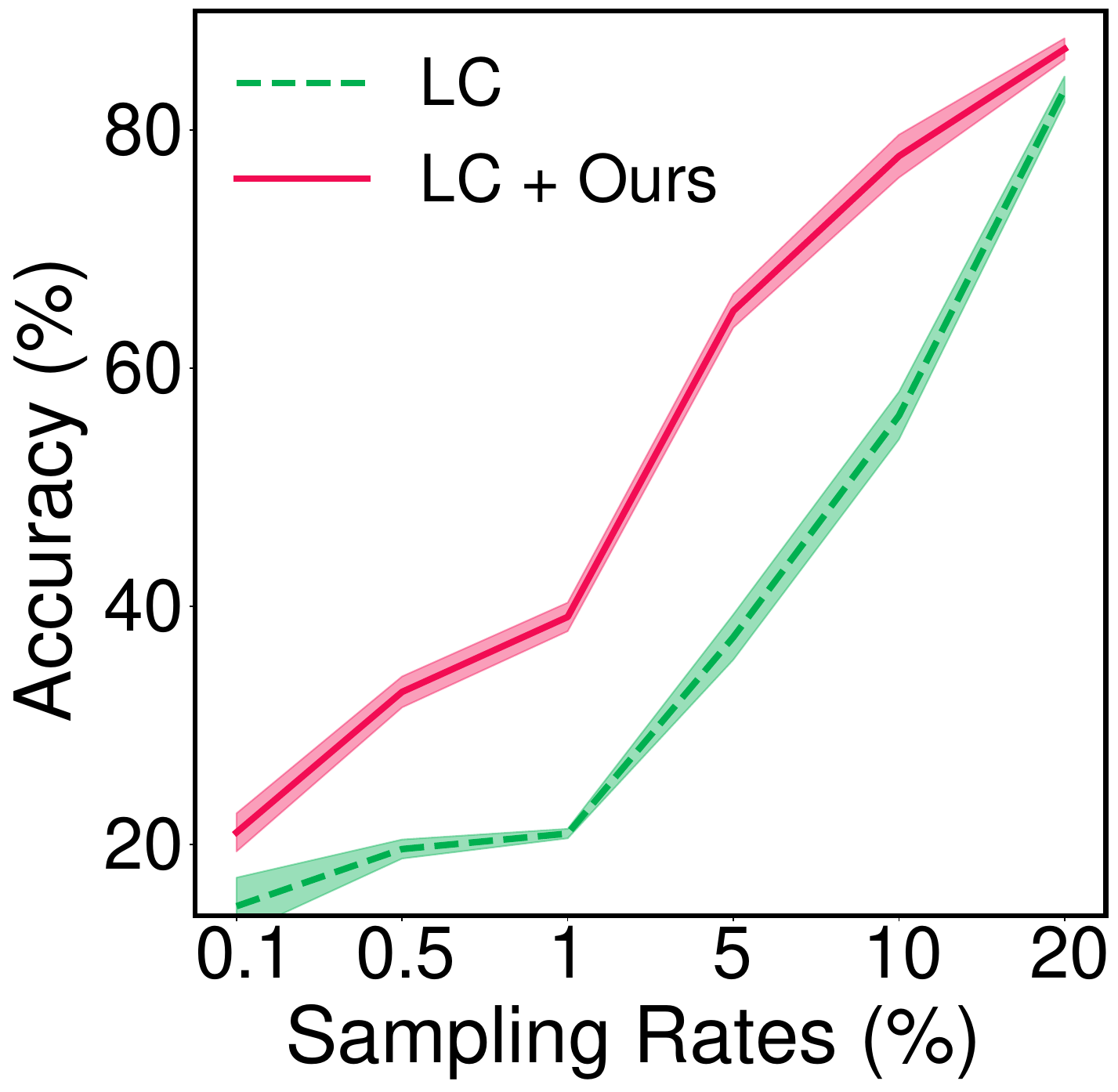}
}
\subfigure[CRAIG]{
\label{fig:f_c}
\includegraphics[width=0.305\columnwidth]{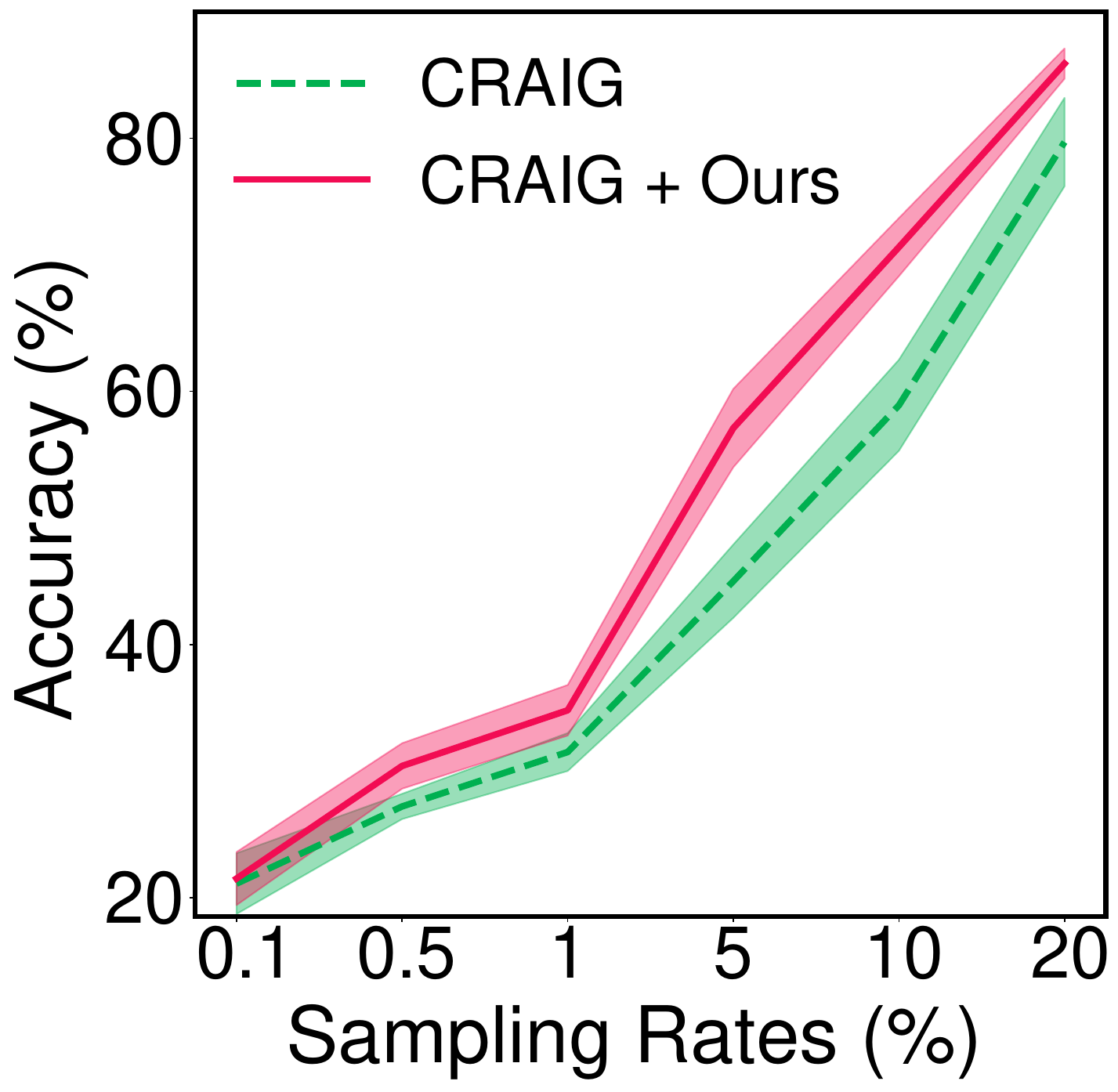}
}
\centering
\subfigure[KCG]{
\label{fig:three_e}
\includegraphics[width=0.305\columnwidth]{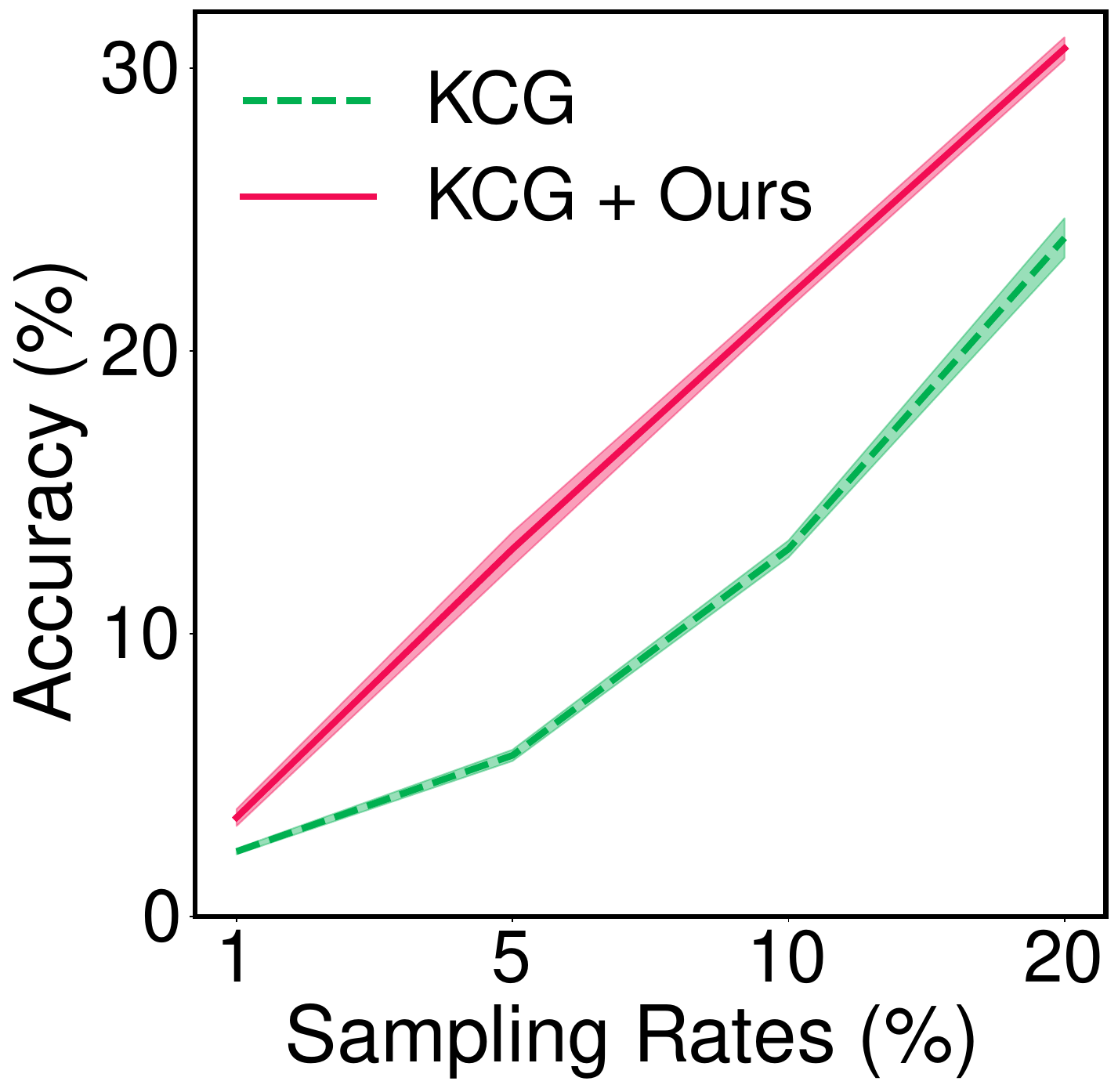}}
\subfigure[LC]{
\label{fig:three_f}
\includegraphics[width=0.305\columnwidth]{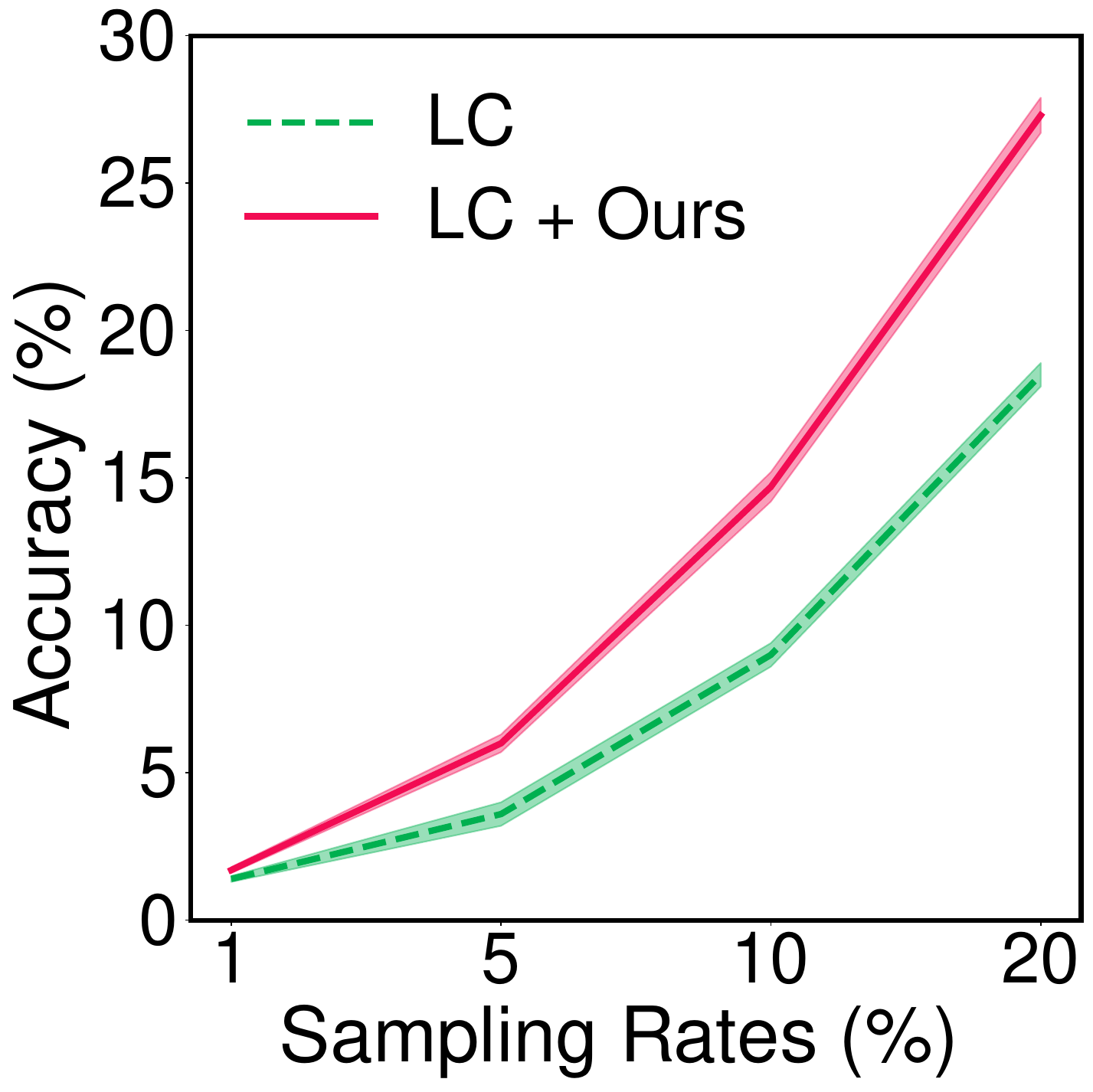}
}
\subfigure[CRAIG]{
\label{fig:three_g}
\includegraphics[width=0.305\columnwidth]{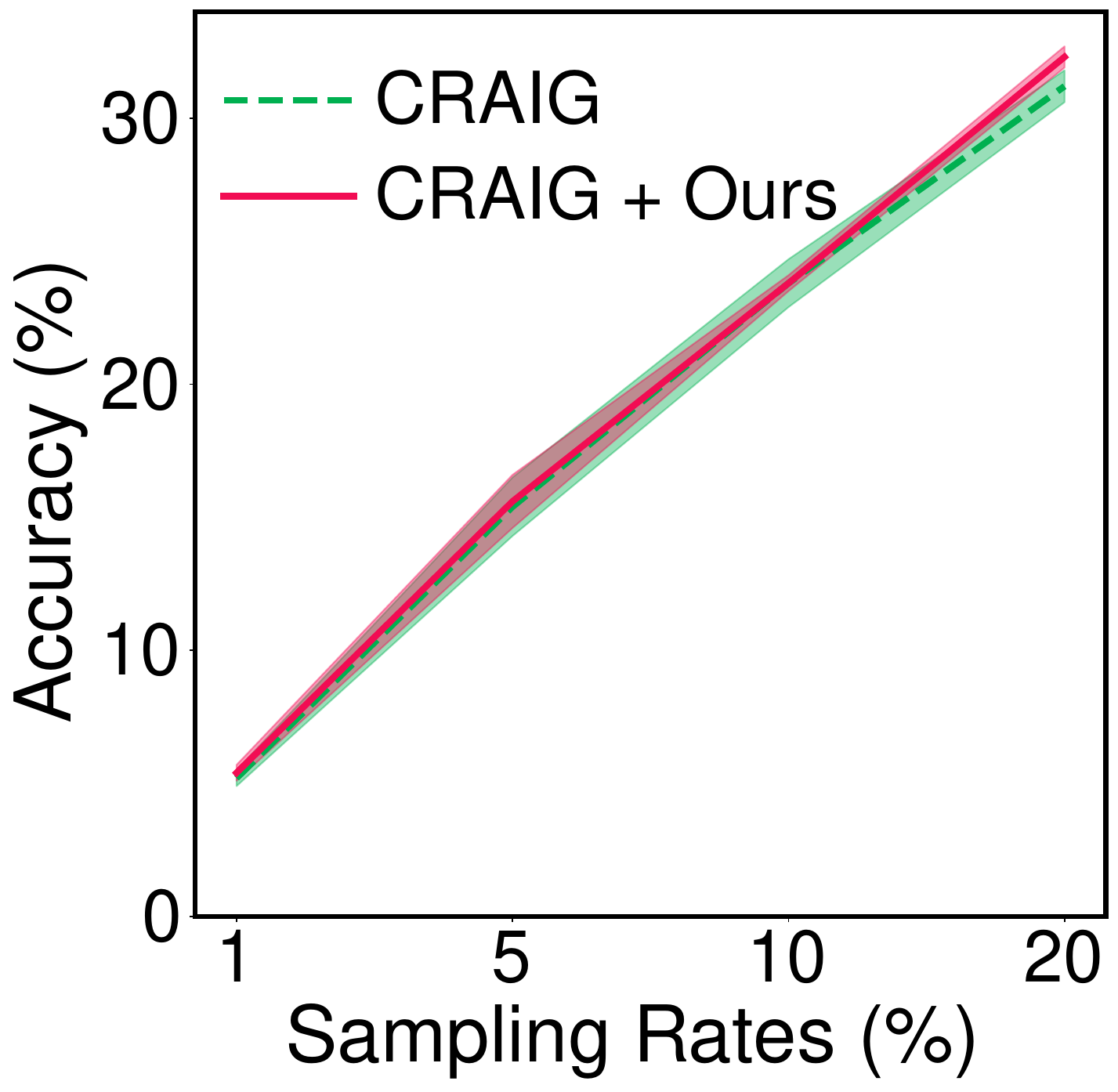}
}
\caption{Performance improvement over baselines. We improve current methods with our proposed CDS metric and constraint. We compare the improved versions with respective baselines on CIFAR-10 (a--c) and TinyImageNet (d--f) under the class-balanced sampling setting.
}
\label{fig:comparison_LC}
\end{figure}

\subsection{Results of Class-balanced Sampling}
Table~{\ref{tab:CDS_CIFAR10}} and Figure~{\ref{fig:comparison_LC}} show the experimental results on CIFAR-10 and TinyImageNet. 
Please refer to \emph{{Supplementary Material \ref{sec_results_cifar100}}} for the results for CIFAR-100. We select subsets of CIFAR-10 with fractions of 0.1\%, 0.5\%, 1\%, 5\%, 10\%, 20\% of the full training dataset respectively. For the best experiments of methods KCG, CRAIG, GC and M-DS on the CIFAR-10, we use the PCA algorithm to select the 10 most relevant dimensions of the features extracted from the network for the CDS metric and set $\beta$ = 1e-4; for the best experiments of LC on the CIFAR-10, we directly use the 512-dimensional features extracted from the network for the CDS metric and set $\beta$ = 1e-3. To ensure that a minimum of 5 images are sampled from each class of TIN dataset, we start with a 1\% sampling rate and select subsets of the TIN with fractions of 1\%, 5\%, 10\%, 20\% of the full training dataset. We select the 10 least relevant dimensions for the best experiments of all baselines on TIN, with $\beta =$ 1e-1.

In Table~{\ref{tab:CDS_CIFAR10}}, we first report the improved results based on two baselines with optimal performance at different sampling rates, \ie GC and M-DS. When comparing the baseline+Ours to the baseline, it is observed that our method enhances the performance of baselines at all sampling rates. When further comparing baseline+Ours with other coreset selection methods, it can be seen that our method further strengthens the leading power of the GC at the 0.1\%-5\% sampling rates, and even makes the overall performance of M-DS better than the Random method on CIFAR-10. The results prove the effectiveness of our method.

In Figure~\ref{fig:comparison_LC}, we show the improved results of baselines KCG, LC, CRAIG on CIFAR-10 and TIN datasets. The results consistently show that our method effectively improves these three types of coreset selection methods, proving the general effectiveness of our method. In particular, taking the LC as an example, LC+Ours outperforms LC by an average of 15.0\% accuracy when sampling 0.1\% to 20\% of CIFAR-10, while LC+Ours outperforms LC by an average of 4.3\% accuracy when sampling 1\% to 20\% of TIN. 

It needs to be emphasized that KCG, CRAIG, GC, and M-DS employ the overall similarity metric in the measurement stage, whereas LC uses the predicted probability directly without using the overall similarity metric. This means that our method not only improves baselines that use the overall similarity metric but is also effective for other baselines. We reveal by visualizing the distributions that existing methods do not handle subset diversity well, while our method motivates them to capture diversity adequately, thus boosting model performance remarkably well. For length reasons, the visualisations are shown in the \emph{Supplementary Material \ref{sec:tsne}}. 

\begin{figure}[t]
\centering
\subfigure[CIFAR-10]{
\label{fig4_1}
\includegraphics[width=0.47\columnwidth]{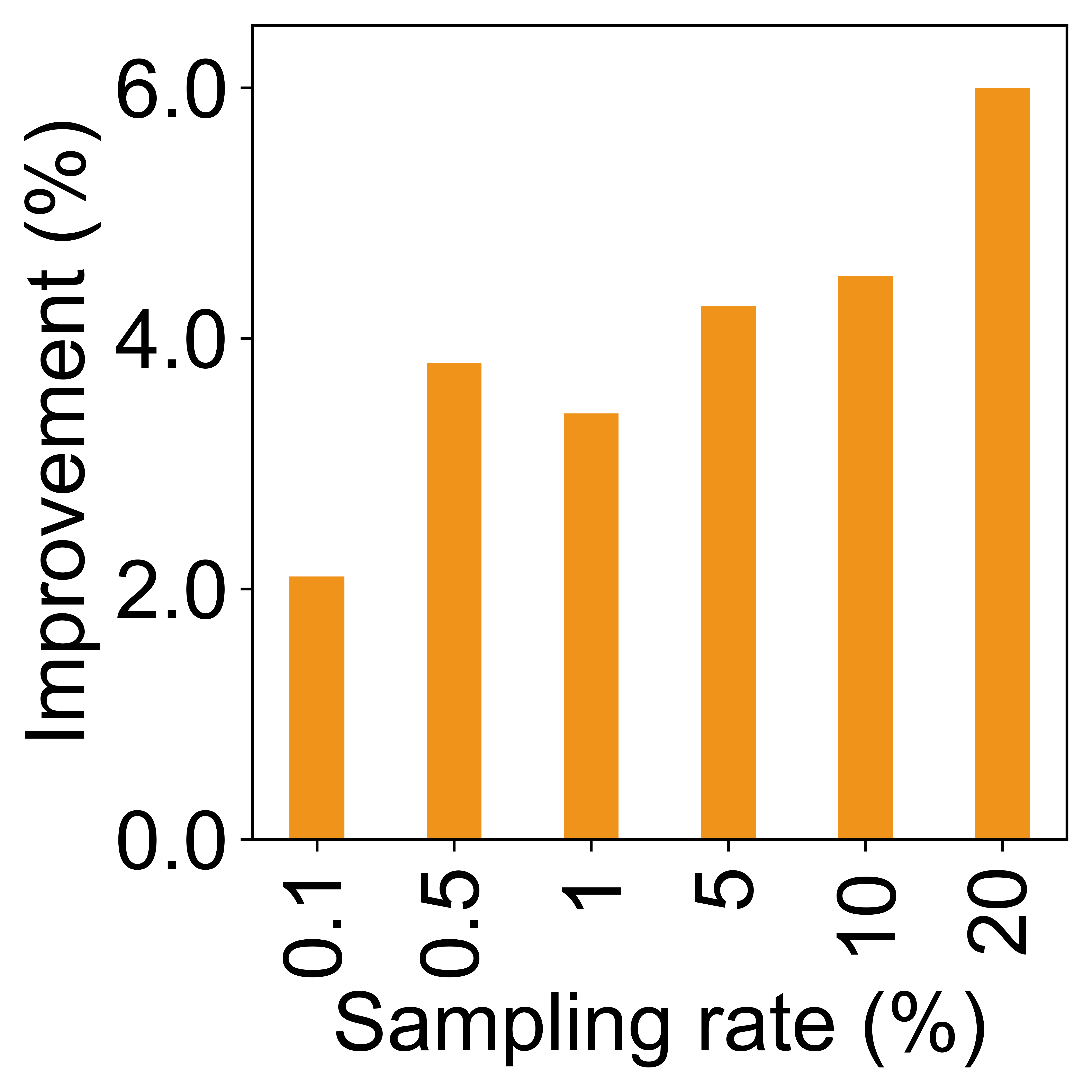}}\subfigure[CIFAR-100]{
\label{fig4_2}
\includegraphics[width=0.47\columnwidth]{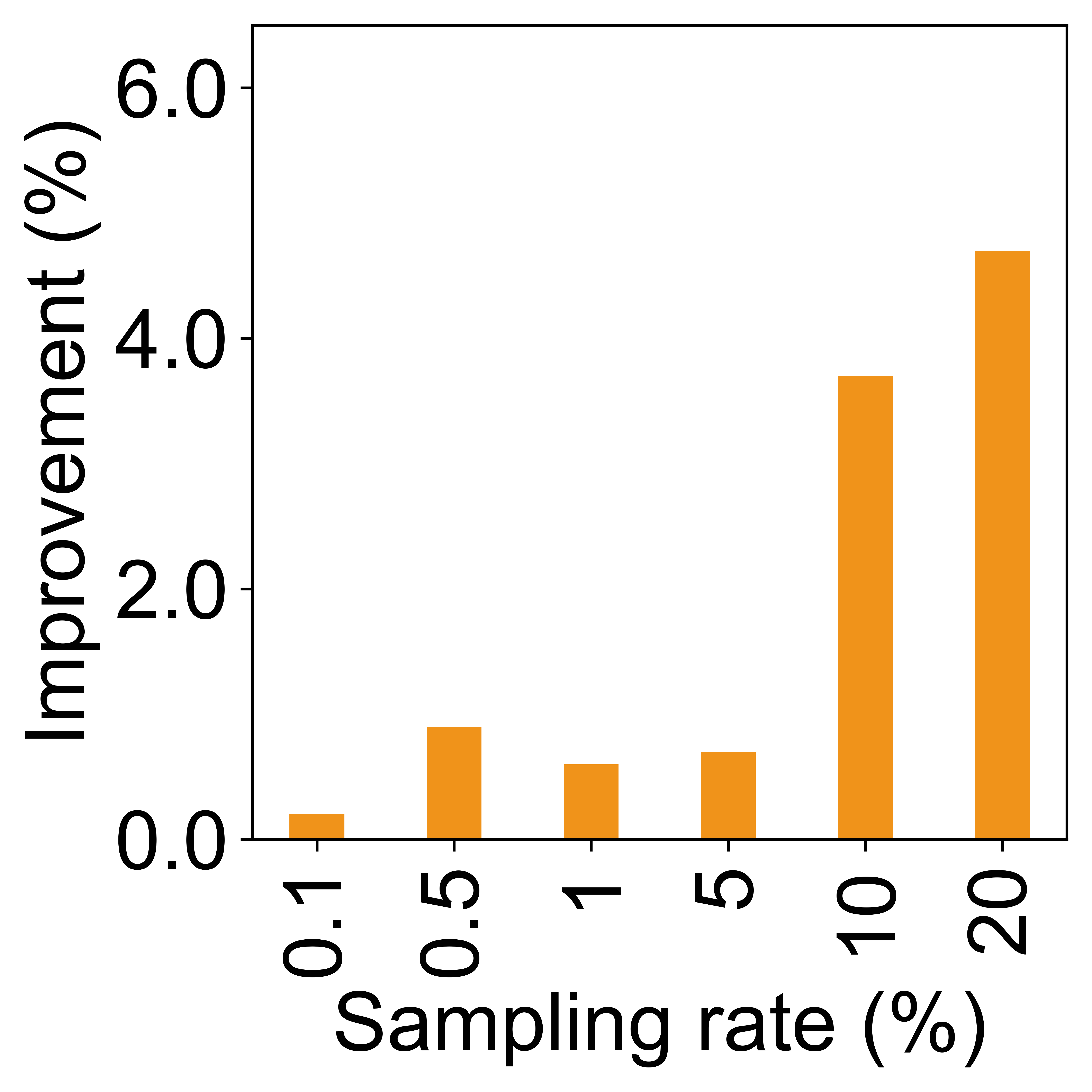}
}
\caption{Performance improvement of GC+Soft CDS over GC in the class imbalanced way
}
\label{fig:ap_imbalance}
\end{figure}

\begin{table}
    \centering
    \small
    \begin{tabular}{cccccc}
    \toprule
         & \rotatebox{0}{\makecell{dim\\reduction}} & \rotatebox{0}{\makecell{partition}} & \rotatebox{0}{\makecell{CDS-r}} &  \rotatebox{0}{\makecell{cons-\\traint}}\\\midrule
        (v1) & \xmark & \xmark & \xmark & \xmark & 34.0$\pm$1.3 \\
        (v2)  & \xmark & \xmark & \xmark & \cmark&  32.8$\pm$0.7 \\
        (v3)  & \cmark& \xmark & \xmark & \cmark&  34.3$\pm$2.5\\
        (v4)& \cmark& \cmark& \xmark & \cmark&  33.5$\pm$1.1\\
        full  &  \cmark & \cmark & \cmark & \cmark & 36.4$\pm$1.0\\
    \bottomrule
    \end{tabular}
    \caption{Ablation study on 0.5\% of the CIFAR-10}
    \label{tab:ablation}
\end{table}

\subsection{Results of Class-imbalanced Sampling}
In this subsection, we choose the method GC to perform class imbalanced sampling on CIFAR-10 and CIFAR-100. Soft CDS is used to improve it. For the experiments on the CIFAR-10, we use the PCA algorithm to select the 10 most relevant dimensions of the features extracted from the network for the CDS metric and set $\beta$ = 1e-4. We select the 10 least relevant dimensions for the experiments on CIFAR-100, with $\beta$ = 1e-2. We show the performance improvement of GC+Soft CDS over GC in Figure~{\ref{fig:ap_imbalance}}. It can be seen that soft CDS effectively improves the performance of GC on both datasets. For example, Soft CDS improves the performance of GC by an average of 4.0\% when sampling 0.1\% to 20\% of the CIFAR-10.

\subsection{Ablation and Parameter Studies}
\label{ablation}
\paragraph{Ablation Study} 
Our method consists of four parts: dimension reduction (dim. reduction), partition, CDS relationship (CDS-r), and CDS diversity constraint (constraint). Since the validity of the constraint has been demonstrated in Figure~{\ref{fig:analysis}(a)}, we evaluate the effectiveness of the other three parts based on the GC (denoted as v1) in Table~{\ref{tab:ablation}}. When introducing the feature into the GC using $L$2-norm and constraining the CDS diversity based on the feature similarity (as v2), it gets a performance drop of 1.2\%. When the dim. reduction part is added (as v3), it obtains a 1.5\% accuracy improvement, exceeding the performance of baseline GC. When the partition part (as v4) is then introduced, the performance is harmed. However, when $L$2-norm is replaced with CDS metric to compute CDS relationship for CDS constraint (as full), the performance is optimal, \ie achieving 36.4$\small\pm1.0$ accuracy. It can prove the effectiveness of dim. reduction and CDS-r, while the partition is valid in the CDS metric.

\begin{figure}
\centering
\subfigure[$k$]{
\label{pa_1}
\includegraphics[width=0.47\columnwidth]{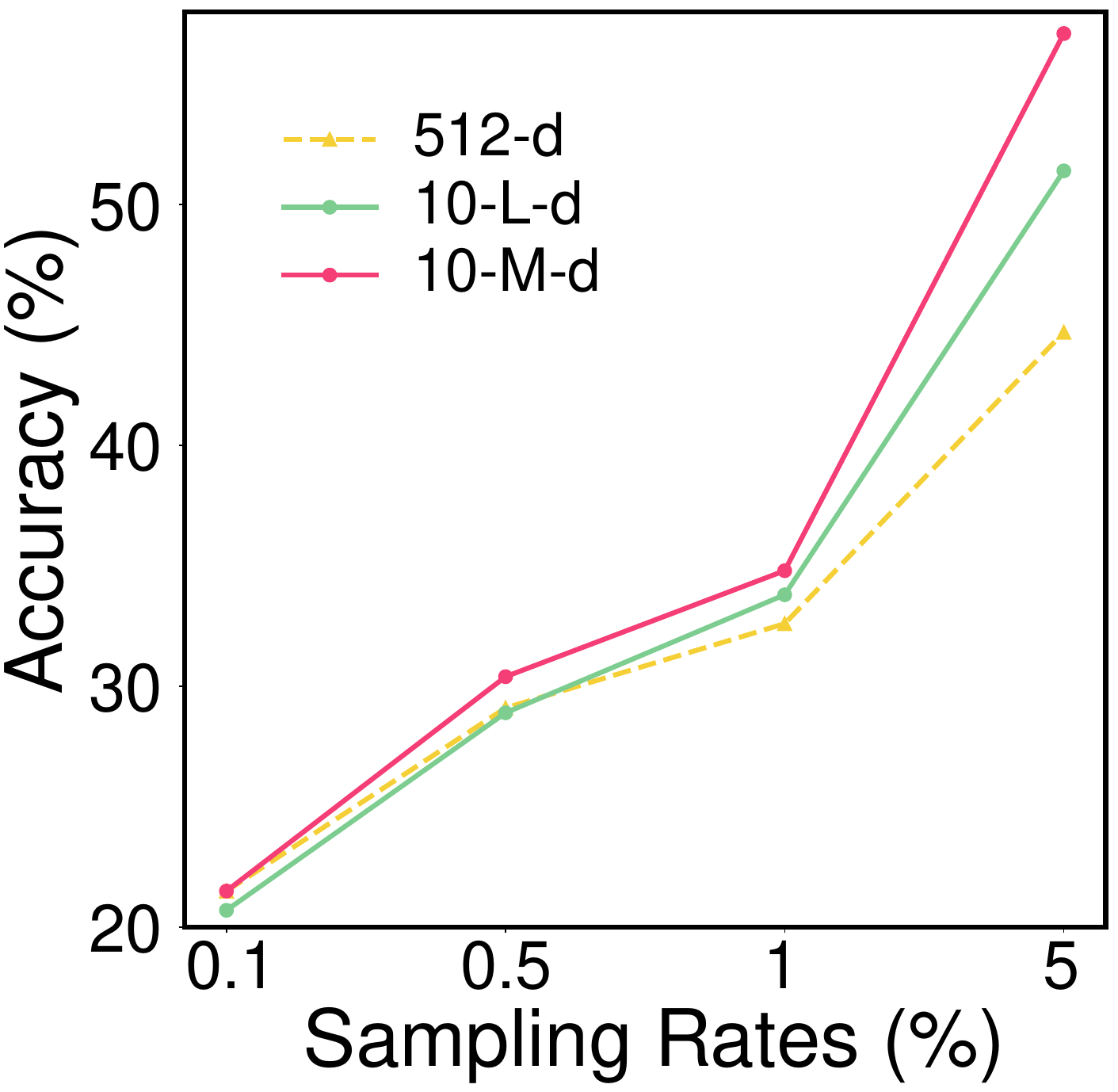}}\subfigure[$\beta$]{
\label{pa_2}
\includegraphics[width=0.47\columnwidth]{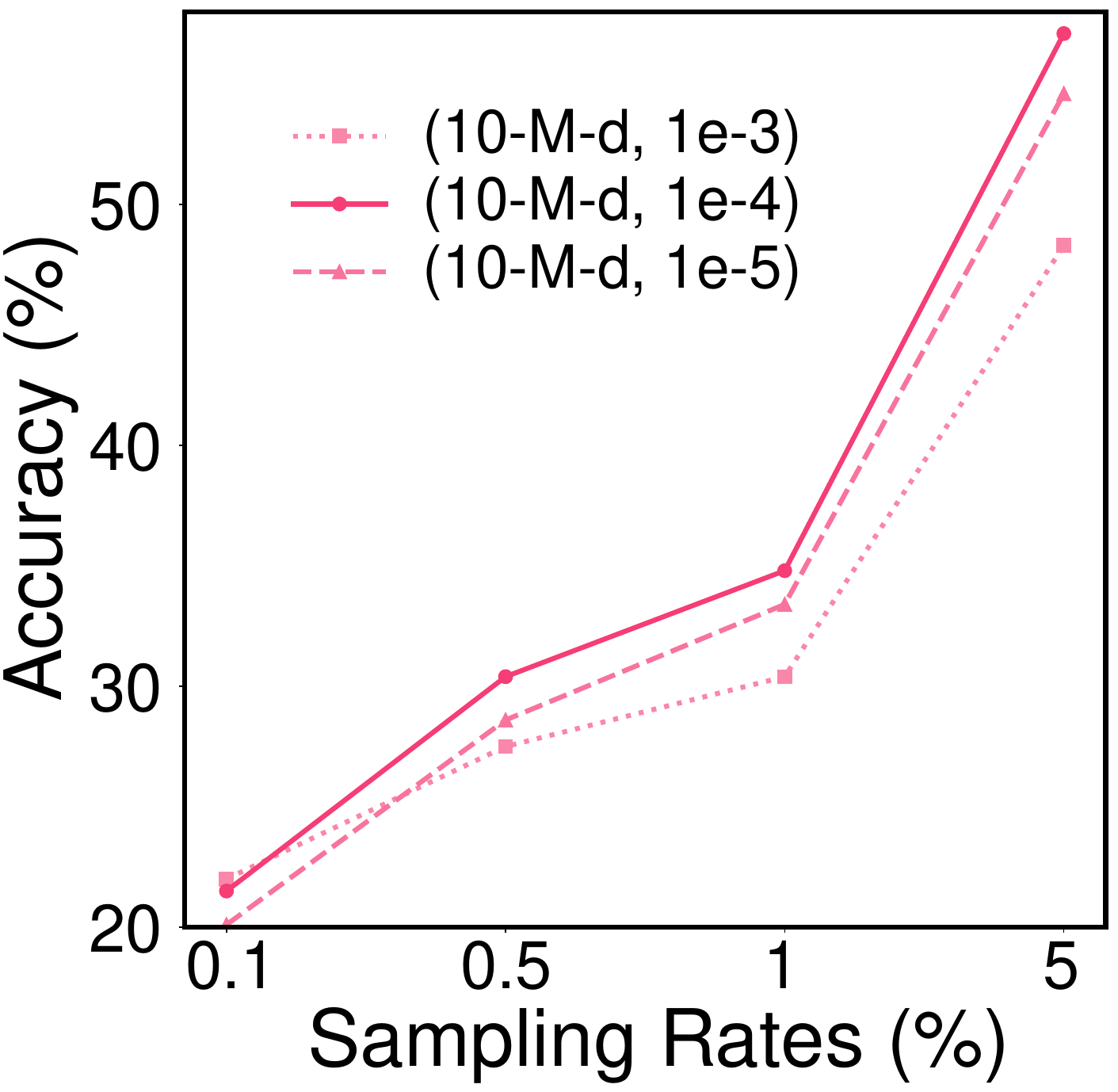}
}
\caption{Parameter analysis. It shows that our method achieves the best improvement compared to the baseline method (CRAIG) when $K=$10-M-D and $\beta=$1e-4. 
}
\label{fig:k_b}
\end{figure}

\paragraph{Parameter Study} The dimension $k$ of the pruned feature and the contribution threshold $\beta$ are important parameters for our method. We first study the effect of $k$ in Figure~{\ref{fig:k_b}(a)}, and then study the effect of $\beta$ based on the best choice of $k$ in Figure~{\ref{fig:k_b}(b)}. We have tried three kinds of $k$, namely (1) original feature extracted before the final fully connected layer, where $k=512$-D; (2) the 10 most relevant dimensions (10-M-D) of the feature extracted from the network; (3) the 10 least relevant dimensions (10-L-D) of the feature extracted from the network. Each type of $k$ corresponds to one $\tilde{\beta}$, which is the maximum value that satisfies $\sum_{i \in n}\sum_{j \in k}\Theta_{ij} / (n\times k) \geq 0.9$. Based on the optimal $k$, we empirically set $\beta \in \{10 \times \tilde{\beta}, \tilde{\beta}, 0.1 \times \tilde{\beta} \}$ to find the optimal $\beta$. 

\section{Conclusion}
This paper introduces CDS to the coreset selection and a novel CDS metric for evaluating diversity. Utilizing this metric, we propose a CDS constraint to augment diversity within coreset selection methods. Our extensive experimental results affirm the effectiveness of our approach across a spectrum of methods. The current pipeline does not take the time cost of selecting data into account. In future work, we hope to design a more efficient and robust baseline for coreset selection.

\section*{Acknowledgements}
We thank our anonymous reviewers for valuable feedback. This research was supported by Hubei Key R\&D (2022BAA033), National Natural Science Foundation of China (62171325), A*STAR AME Programmatic Funding A18A2b0046, RobotHTPO Seed Fund under Project C211518008, EDB Space Technology Development Grant under Project S22-19016-STDP, JSPS KAKENHI JP22H03620, JP22H05015, and the Value Exchange Engineering, a joint research project between Mercari, Inc. and RIISE. The Supercomputing Center of Wuhan University supports the supercomputing resource.

\bibliography{aaai24}

\appendix

\clearpage

\twocolumn[
\begin{@twocolumnfalse}
\section*{\centering{Supplementary Material for \emph{Contributing Dimension Structure of\\ Deep Feature for Coreset Selection\\[25pt]}}}
\end{@twocolumnfalse}
]

\begin{table*}[t]
\centering
\begin{tabular}{c|l}
\toprule
Notations &  Description \\ \hline
$\mathcal{D}$, $n$ & Full training dataset and its size\\ \hline
$\mathcal{S}$, $b$ & Training subset and its size (sampling budget)\\ \hline
$\mathcal{V}$ & Unselected data, $\mathcal{V} = \mathcal{D} \setminus \mathcal{S}$ \\ \hline
$x_{i}$, $y_{i}$ & A training sample and its class label \\ \hline
$C$ & The number of classes, $\in \mathbb{R}$ \\ \hline
$\theta_{pre}$, $\theta^\mathcal{D}$, $\theta^\mathcal{S}$ & The pre-trained model, the model trained on $\mathcal{D}$ and the model trained on $\mathcal{S}$ \\ \hline
${F}(x_{i})$, $f_{i}^j$ & The feature of sample $x_{i}$, and the $j^{th}$ dimension of the feature ${F}(x_{i})$ \\ \hline
$\tilde{{F}}$ & The central feature (of the class or dataset) \\ \hline 
$\Xi(x_{i})$ & The CDS of the sample $x_{i}$\\ \hline
${R}^{c}$ & CDS relationship matrix for class $c$ \\ \hline
$K$, $k$ & \multicolumn{1}{l}{\begin{tabular}[c]{@{}l@{}}The number of dimensions of the deep feature extracted from the model \\and that of the deep feature after dimension reduction, $k \leq K$  \end{tabular} }\\ \hline
$T(\cdot)$ & The objective function proposed in the existing coreset selection method\\ \hline
$d(\cdot, \cdot)$ & $L_2$-norm (also known as Euclidean distance metric)\\ \hline
$\mathbb{I}(\cdot)$ & A binary decision function, $\in \{0, 1\}$\\ \hline
$H(\cdot)$ & Our designed constraint function \\ \hline
$\alpha$ & A hyperparameter used in the Hard CDS constraint for the $1^{st}$ stage clustering \\ \hline
$\beta$ & An important hyperparameter for the CDS metric\\ 
\bottomrule
\end{tabular}
\caption{Main notations.}
\label{notation}
\end{table*}

\section{Summary of Notations}
\label{sec:notations}
We present a summary of the main notations used throughout this paper in Table~\ref{notation}.

\section{Related Work}
In the face of big data, two dominant methods are currently used to reduce computational costs by reducing the size of the training set: dataset distillation and coreset selection. 

To reduce the size of the training set without losing too much performance, dataset distillation~\cite{cazenavette2023generalizing} uses data generation techniques to encapsulate the knowledge from the full training set into a small number of synthetic training images. It is not limited to the sample distribution in the original dataset when synthesizing data, and does not pursue the realism of the generated data. However, dataset distillation faces challenges such as high optimization costs and difficulty when applied to large-scale high-resolution datasets. 

In this paper, we specifically concentrate on coreset selection and its potential for data-efficient learning, distinct from the method of dataset distillation. Coreset selection is a long-standing learning problem that aims to select a subset of the most informative training samples from the full training set to facilitate data-efficient learning~\cite{guo2022deepcore, das2021tmcoss}. Initially, early coreset selection methods were tailored to accelerate the learning and clustering for machine algorithms like k-means and k-medians~\cite{har2007smaller}, support vector machines~\cite{tsang2005core}, and Bayesian inference~\cite{campbell2018bayesian}. However, these methods are designed for particular models and problems, and have limited applications in deep learning. Recently, with the rapid development of deep learning, research on coreset selection for deep learning has emerged, including geometry-based methods~\cite{sener2017active, agarwal2020contextual}, uncertainty-based methods~\cite{coleman2019selection}, submodularity-based methods~\cite{iyer2021submodular, kothawade2022prism, rangwani2021s3vaada}, gradient-matching-based methods~\cite{mirzasoleiman2020coresets, killamsetty2021grad}, and others~\cite{toneva2018empirical, swayamdipta2020dataset, paul2021deep}. They typically rely on a pre-trained model to obtain information, \eg features, gradients, predicted probabilities, and loss, for measuring the importance of data in the full training set through designed constraints. The importance of data in coreset selection is assessed in two main aspects: representation and diversity. Representation is guaranteed through the distribution matching between subsets and the full set, under the guidance of similarities between data in features, gradients, or other information. Meanwhile, diversity is assured through the imposition of penalties upon similar data. Information similarity is commonly computed by similarity metrics, such as $L$2-norm and cosine distance. However, these similarity metrics obtain the final similarity by simply aggregating dimension similarities without evaluating the impact of different dimensions. This risks treating some distinct samples as equally important, leading to an ineffective assessment of the diversity. Also, non-similarity metric-based methods do not consider this difference and inevitably fall short of effectively capturing diversity. To solve this problem, we propose a feature-based diversity constraint, compelling the chosen subset to exhibit maximum diversity. 

\section{ More Experimental Details and Additional Results}

\subsection{Additional Datasets Details}
\label{sec:datasets_details}
We performed experiments on three common image classification datasets, including CIFAR-10~\cite{krizhevsky2009learning}, CIFAR-100~\cite{krizhevsky2009learning}, and TinyImageNet. CIFAR-10 consists of 60,000 images of $32 \times 32 \times 3$ size in 10 classes, with 6,000 images per class. The training and test sets contain 50,000 and 10,000 images, respectively. Compared to the CIFAR-10, the CIFAR-100 is more fine-grained. It has 100 classes, each with 500 training images and 100 test images. TinyImageNet, a subset of ImageNet~\cite{krizhevsky2012imagenet}, consists of 100,000 training images in 200 classes and 10,000 testing images. In Table~\ref{tab:dataset_info}, we summarize the number of classes, and the number of instances in each split in the image datasets.

\begin{table}[t]
\centering
\begin{tabular}{c|c|c|c}
\toprule
Datasets &  \#Classes & \#Train & \#Test \\ \hline
CIFAR-10 & 10 & 50,000 & 10,000 \\ \hline
CIFAR-100 & 100 & 50,000 & 10,000 \\ \hline
TinyImageNet & 200 & 100,000 & 10,000 \\
\bottomrule
\end{tabular}
\caption{Number of classes, number of instances in Train and Test split in each image datasets.}
\label{tab:dataset_info}
\end{table}

\subsection{Experimental Details on More S-CDS and More D-CDS}
\label{sec:details_moreSD}
This section provides details of experimental studies on more S-CDS and more D-CDS, as well as detailed algorithms for more S-CDS, more D-CDS, and more Random.

\heading{Dataset, model and experimental setup} To discuss and compare the role of more D-CDS and more S-CDS strategies for coreset selection, we perform experiments on the common image classification dataset, \ie CIFAR-10~\cite{krizhevsky2009learning}.

\SetKw{KwBy}{by}
\begin{algorithm}[t]
    \small
    \caption{ Coreset Selection with Hard CDS Constraint (Class Balanced Sampling)}
    \label{alg:algorithm_hard}
    \SetAlgoLined
    \KwIn{Train set: $\mathcal{D}$; data classes: $C$; budget: $b$; pre-trained model: $\theta_{pre}$; objective function: $T$; parameter $\alpha$; parameter $\beta$.}
    \KwOut{Coreset $\mathcal{S} \subset \mathcal{D}$.}
    Initial $\mathcal{S} \leftarrow \mathcal{S}_{0}$\\
    \For{$i\gets0$ \KwTo $C-1$ \KwBy $1$}{
    ${F} \leftarrow M(\mathcal{D}^{i}; \theta_{pre})$\\
    ${F} \leftarrow \PCA({F})$\\
    $\tilde{{F}} \leftarrow \mean({F})$\\
    \# $1^{st}$ stage clustering: obtain $L$ data clusters\\
    \For{$j\gets0$ \KwTo $n_{i}-1$ \KwBy $1$}{
    $d_{jc} \leftarrow d({F}_{j}, \tilde{{F}})$\\
    $l \leftarrow d_{jc} \mid {\alpha}$\\
    $Cluster(l) \leftarrow Cluster(l) \cup {F}_j $\\
    }
    \# $2^{nd}$ stage clustering\\
    \For{$j\gets0$ \KwTo $L-1$ \KwBy $1$}{
    $b_{ij} \leftarrow b \times \lfloor n_{ij}/n_{i} \rfloor$\\
    ${R}^{j} \leftarrow $ CDS-Metric$({Cluster(j)}, \beta)$\\
    $t, b_{t} \leftarrow $ Cluster-Hard-CDS($Cluster(j),{R}^{j}, b_{ij}$)\\
    \# Data selection\\
    Initial $\mathcal{S}^{'} \leftarrow \mathcal{S}_{0}$\\
    \For{$m\gets0$ \KwTo $|t|-1$ \KwBy $1$}{
        $s \leftarrow \argmax_{s \subset t_{m}}T\left(s\right) \quad s.t. \quad \left|s\right| \leq b_{m}$\\
        $\mathcal{S}^{'} \leftarrow \mathcal{S}^{'} \cup \{s\}$\\
        }
    $\mathcal{S} \leftarrow \mathcal{S}^{'}$\\
    }
}
\end{algorithm}

We utilize the 18-layer residual network (ResNet-18)~\cite{he2016deep} as the backbone of the pre-trained model and target model, and use the deep features extracted before the final fully connected layer for the CDS metric. The deep feature is 512 dimensions (512-D) in this paper. We follow the setup of work~\cite{guo2022deepcore}. Specifically, we use SGD as the optimizer with batch size 128, initial learning rate 0.1, Cosine decay scheduler, momentum 0.9, weight decay $5 \times 10^{-4}$, 10 pre-trained epochs and 200 training epochs. We apply random crop with 4-pixel padding and random flipping on the $32\times32$ training images for data augmentation. For each algorithm, $\alpha=0.5, \beta=$1e-4. As an evaluation metric, we use the classification accuracy. Since the class label of each data is known, to avoid the problem of long-tailed distribution caused by sampling, here we assume a class-balanced sampling setting. We repeat the same experiment 5 times with random seeds for each selection method, and report the performance mean. All experiments were run on Nvidia Tesla V100 GPUs.

\heading{Algorithm of More S-CDS} It constrains as much data in the subset sampled as possible to have the same CDS. This means sampling a budget-sized subset with fewer types of CDS. Similar to Algorithm~\ref{alg:algorithm_hard}, the coreset selection with more S-CDS performs a two-stage clustering and a data selection. The differences are that (1) more S-CDS directly uses the 512-dimensional features extracted from the pre-trained model for the calculation of feature distance and the CDS metric; (2) in addition, after subdividing $Cluster(j)$ into multiple clusters $t$ according to ${R}^{j}$, the clusters $t$ are sorted according to their size from largest to smallest, with the expectation that the budget will be allocated to the largest Cluster only. If the budget is larger than the size of the largest Cluster, the excess budget is allocated to the second largest Cluster, and so on, until the budget is allocated; (3) in the data selection stage, samples of budget size are directly selected from each cluster $t$ using the Random method.\\

\heading{Algorithm of More D-CDS} It constrains the subset sampled to have as many different CDS as possible. This means sampling a budget-sized subset with more types of CDS. Similar to Algorithm~\ref{alg:algorithm_hard}, the coreset selection with more D-CDS performs a two-stage clustering and a data selection. The differences are that (1) more D-CDS directly uses the 512-dimensional features extracted from the pre-trained model for the calculation of feature distance and the CDS metric; (2) in the data selection stage, samples of budget size are directly selected from each cluster $t$ using the Random method.\\

\heading{Algorithm of More Random} Firstly, it performs the similar $1^{st}$ stage clustering as the Algorithm~\ref{alg:algorithm_hard}. Differently, it directly uses the 512-dimensional features extracted from the pre-trained model to calculate feature distance. Then, it selects samples of budget size from $Cluster(j)$ using the Random method.

\subsection{Algorithm of Coreset Selection with Hard CDS}
\label{sec:cs_hard CDS}
The coreset selection algorithm with Hard CDS\footnote{The code is available in https://github.com/Vivian-wzj/contributing-dimension-structure} consists of a two-stage clustering process and a data selection process. To describe it more clearly, we cover each step in detail in Algorithm~\ref{alg:algorithm_hard}. The most critical step in implementing the CDS Constraint is in line 16 of Algorithm~\ref{alg:algorithm_hard}, where the function Cluster-Hard-CDS subdivides $Cluster(j)$ into clusters $t$ according to ${R}^{j}$. Samples within clusters have the same CDS, while samples between clusters have different CDSs. After that, the sampling budget $b_{t}$ of each cluster is constrained to be as equal as possible, thus increasing the diversity of the CDS in the coreset. The $b_{t}$ stores the sampling budgets that are allocated to each cluster, and it satisfies sum$(b_{t})$ = $b_{ij}$.

\SetKw{KwBy}{by}
\begin{algorithm}[tb]
    \caption{Coreset Selection with Soft CDS Constraint (Class Balanced Sampling)}
    \label{alg:algorithm_soft}
    \SetAlgoLined
  \KwIn{Train set: $\mathcal{D}$; data classes: $C$; budget: $b$; pre-trained model: $\theta_{pre}$; objective function: $T$; constraint function: $H$; parameter $\beta$.}
  \KwOut{Coreset $\mathcal{S} \subset \mathcal{D}$.}
        Initial $\mathcal{S} \leftarrow \mathcal{S}_{0}$\\
        \For{$i\gets0$ \KwTo $C-1$ \KwBy $1$}{
        $\bm{F} \leftarrow M(\mathcal{D}^{i}; \theta_{pre})$\\
        $\bm{R}^{i} \leftarrow \text{CDS-Metric}(\bm{F}, \beta)$ \\
        Initial $\mathcal{S}' \leftarrow \mathcal{S}_{0}$\\
        \While{$|\mathcal{S}'| < b$}{
        $V \leftarrow \mathcal{D}^{i} \setminus \mathcal{S}'$ \\
        $e \leftarrow \argmax_{e \in V}T\left(e|\mathcal{S}'\right) \times H(\bm{R}^{i})$\\
        $\mathcal{S}' \leftarrow \mathcal{S}' \cup \{e\}$\\
        }
        $\mathcal{S} \leftarrow \mathcal{S}'$\\
        }
\end{algorithm}

\subsection{Results for Hard CDS \textit{vs.} Soft CDS}
\label{sec_hard_vs_soft}
\begin{figure}[t]
\centering
\subfigure[Improvement over CRAIG]{
\includegraphics[width=0.21\textwidth]{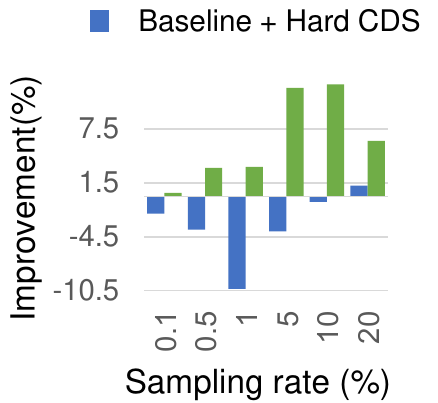}}\subfigure[Improvement over GC]{
\includegraphics[width=0.205\textwidth]{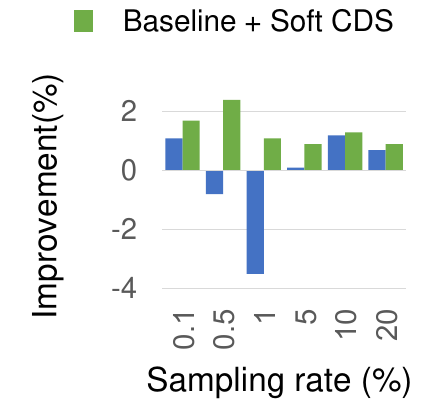}
}
\vspace{-3mm}
\caption{ \textbf{Comparison of Hard CDS and Soft CDS} in improving CRAIG and Graph Cut(GC) performance. We try to improve baselines (\ie CRAIG and GC) with two implementations of the CDS constraint (\ie Hard CDS and Soft CDS) and show their performance improvement over the baseline on the CIFAR-10 dataset. Soft CDS improves baseline performance better than Hard CDS in improving CRAIG and GC.
}
\label{fig:hard_vs_soft}
\vspace{-5mm}
\end{figure}

We report in Figure~\ref{fig:hard_vs_soft} the detailed improvement results on the CIFAR-10 dataset when Hard CDS and Soft CDS are applied to the baselines (CRAIG~\cite{mirzasoleiman2020coresets} and Graph Cut (GC)~\cite{iyer2021submodular}), respectively. It can be seen that:

(1) \textbf{Hard CDS works for most methods}. Although Hard CDS effectively improves the performance of Least Confidence and K-Center Greedy methods on the CIFAR-10 dataset (as shown by the experiments on class-balanced sampling in the main paper), it fails to effectively improve the performance of the CRAIG and GC methods on the CIFAR-10 datasets.

(2) \textbf{Soft CDS, specially designed for the objective function, improves baselines performance better than Hard CDS}. As shown in the Figure~\ref{fig:hard_vs_soft}, Soft CDS successfully improves the performance of target baselines at all kinds of sampling rates. This shows the effectiveness of CDS constraint.

Therefore, to effectively improve coreset selection methods, we apply the Hard CDS to Least Confidence and K-Center Greedy methods, and apply the Soft CDS to CRAIG and Graph Cut methods in our paper.

In summary, both CDS constraint implementations have their advantages and disadvantages. When applying CDS constraint to improve other new coreset selection methods, hard CDS can be tried first. If it fails, then soft CDS, \ie designing effective constraint functions for the new objective function, can be adopted.

\begin{figure}[!t]
\centering
\subfigure[KCG]{
\includegraphics[width=0.2\textwidth]{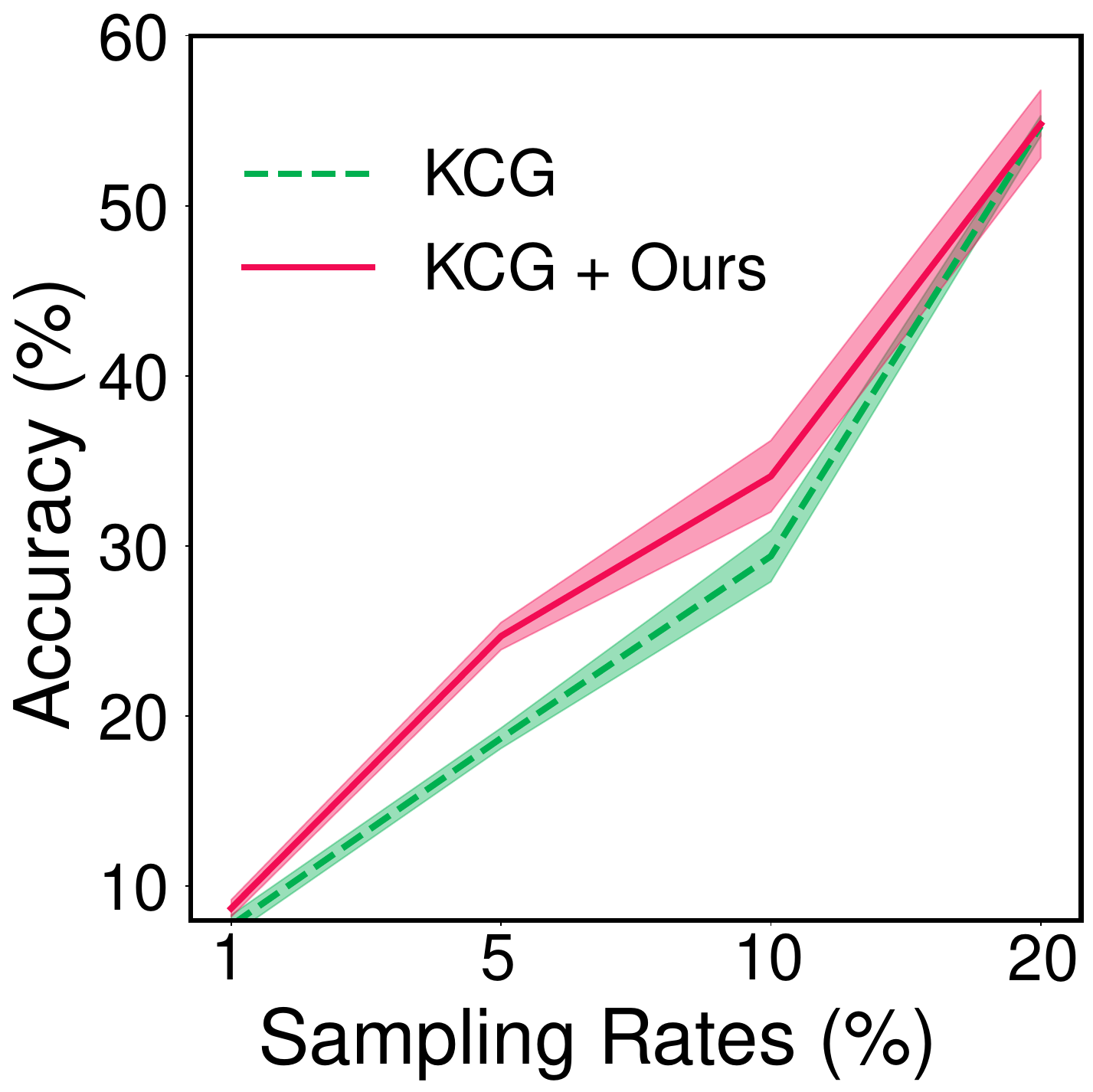}}
\subfigure[LC]{
\includegraphics[width=0.2\textwidth]{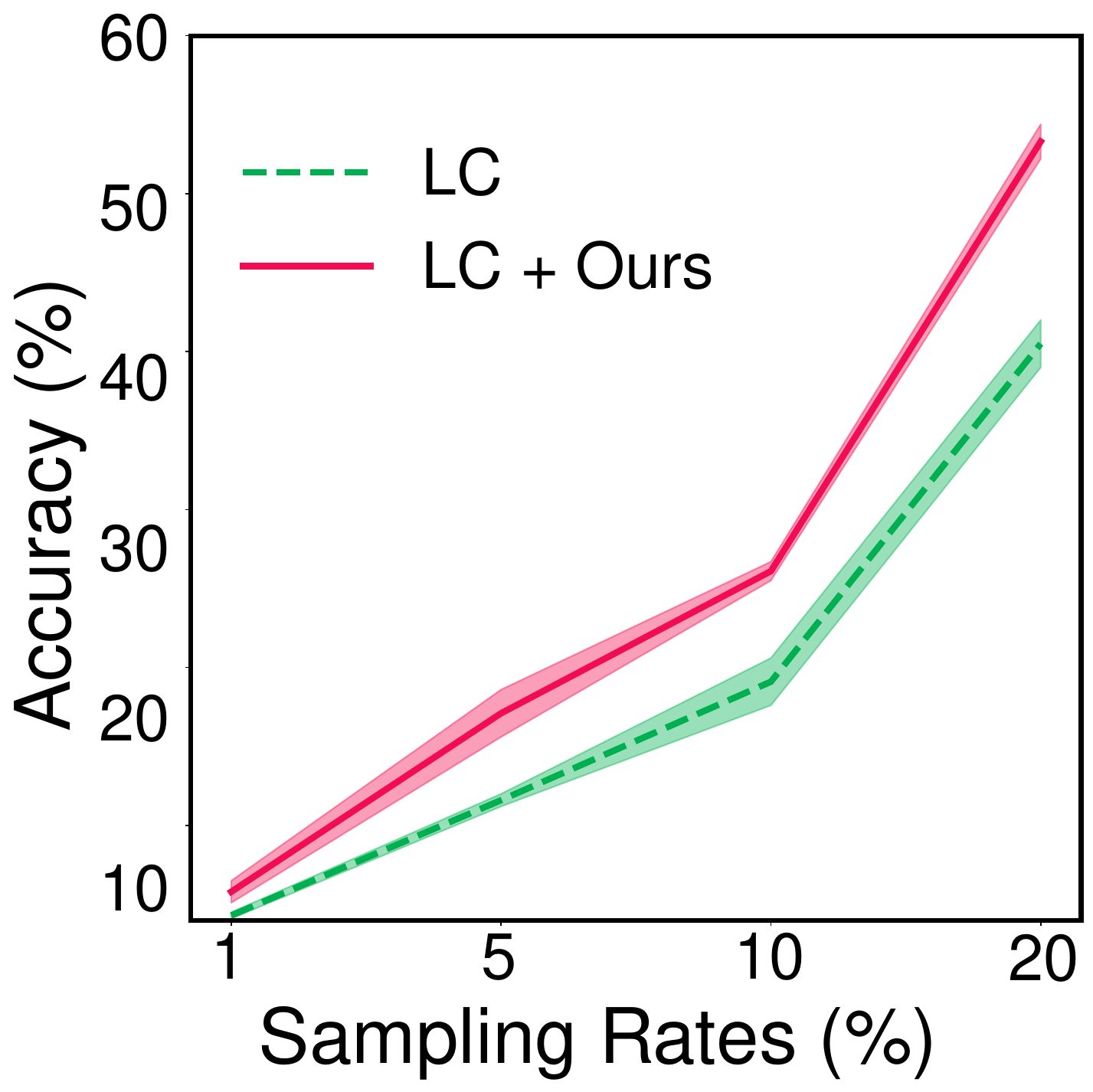}
}

\centering
\subfigure[CRAIG]{
\includegraphics[width=0.2\textwidth]{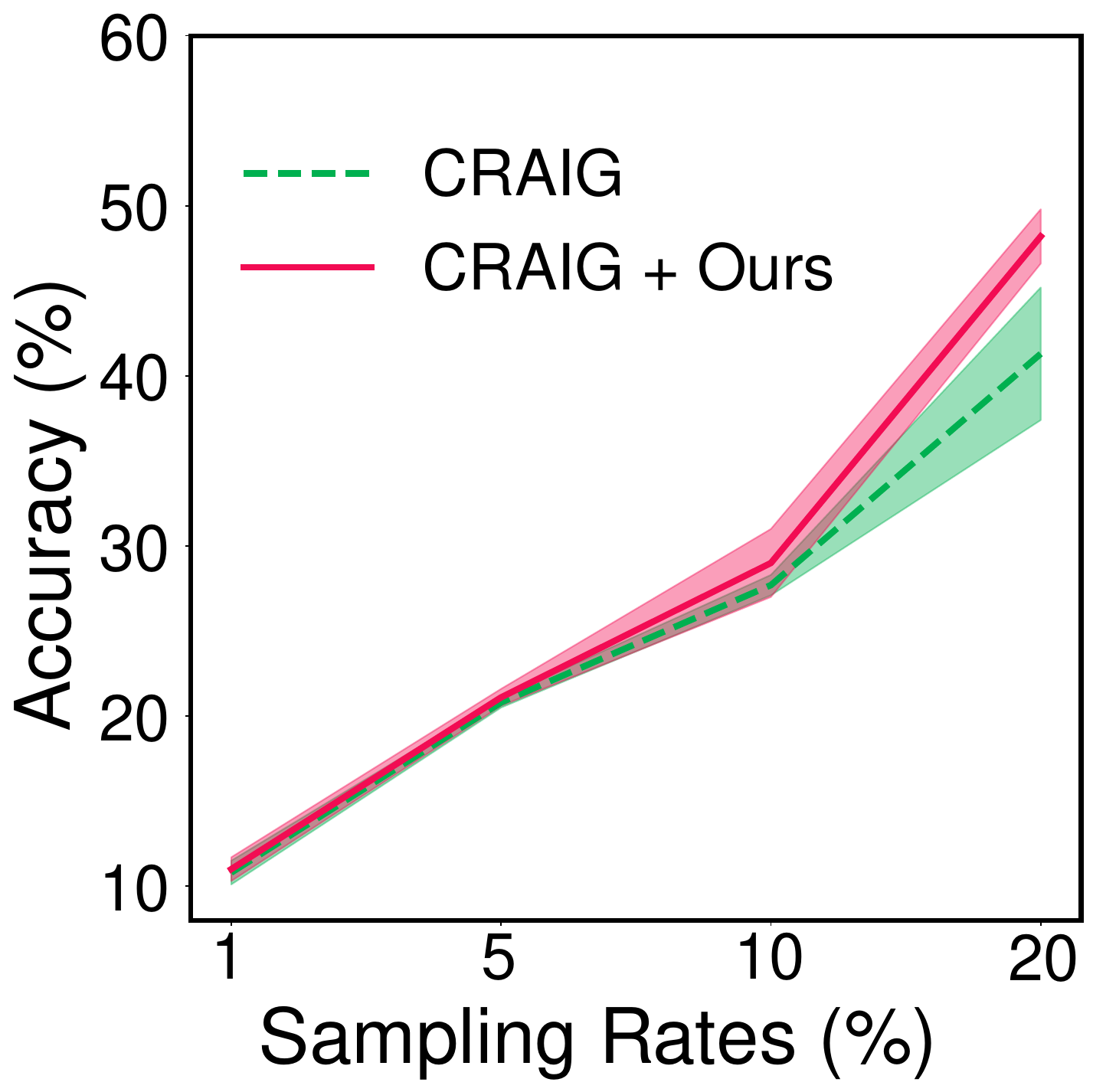}}
\subfigure[GC]{
\includegraphics[width=0.2\textwidth]{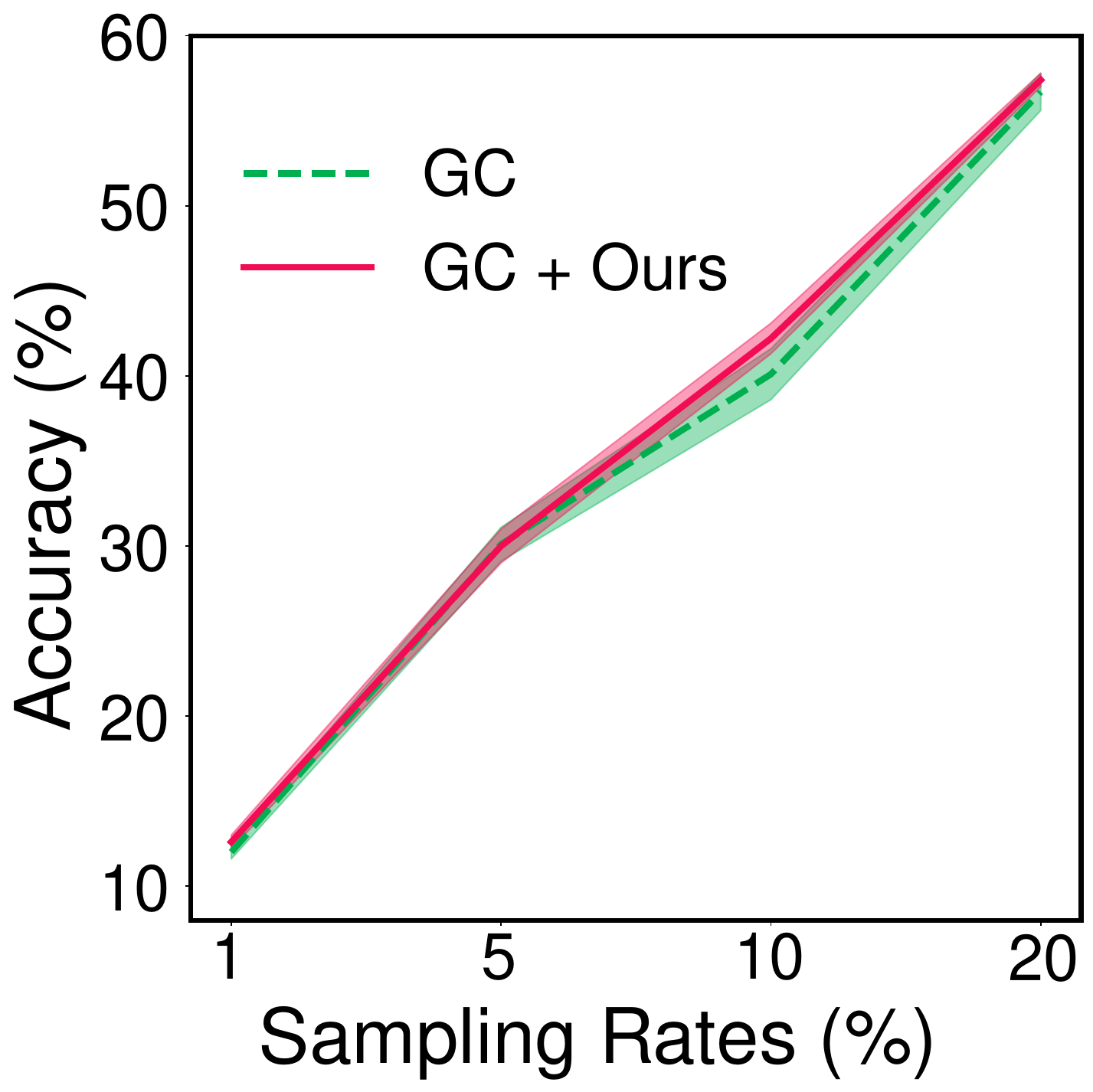}
}
\vspace{-3mm}
\caption{\textbf{Performance improvement over baselines.} We improve baselines with our proposed CDS metric and constraint. We compare the improved versions with respective baselines on CIFAR-100 under the class-balanced sampling setting. The improved versions consistently outperform baselines, suggesting that increasing the diversity of CDS in the coreset can universally enhance existing coreset selection methods.
}
\label{fig:comparison_cifar100}
\vspace{-5mm}
\end{figure}

\subsection{Results on the CIFAR-100 Dataset}
\label{sec_results_cifar100}
We report the effect of our strategy on the CIFAR-100 dataset in Figure~\ref{fig:comparison_cifar100}. To ensure that a minimum of 5 images are sampled from each class of the CIFAR-100 dataset, we start with a 1\% sampling rate and select subsets of the CIFAR-100 dataset with fractions of 1\%, 5\%, 10\%, 20\% of the full training dataset. We select the 10 least relevant dimensions for the best experiments on CIFAR-100, with $\beta =$ 1e-1 for Least Confidence(LC), CRAIG, and GC and $\beta =$ 1e-2 for K-Center Greedy(KCG).

The results in Figure~\ref{fig:comparison_cifar100} consistently show that CDS Constraint can effectively improve the four types of coreset selection methods, proving its general effectiveness. Concretely, when sampling 1\%-20\% of the CIFAR-100 dataset, LC + Hard CDS outperforms LC by an average of 6.7\% accuracy, KCG + Hard CDS outperforms KCG by an average of 3\% accuracy, GRAIG + Soft CDS outperforms GRAIG by an average of 2.2\% accuracy, and GC + Soft CDS outperforms GC by an average of 0.8\% accuracy. This shows the effectiveness of CDS constraint.

Our designed constraint function for each target baseline consistently improves the performance of the baseline at CIFAR-10, CIFAR-100, and TinyImageNet datasets, indicating that the CDS constraint is effective and not limited by the dataset.

\begin{figure}
    \centering
    \includegraphics[width=\columnwidth]{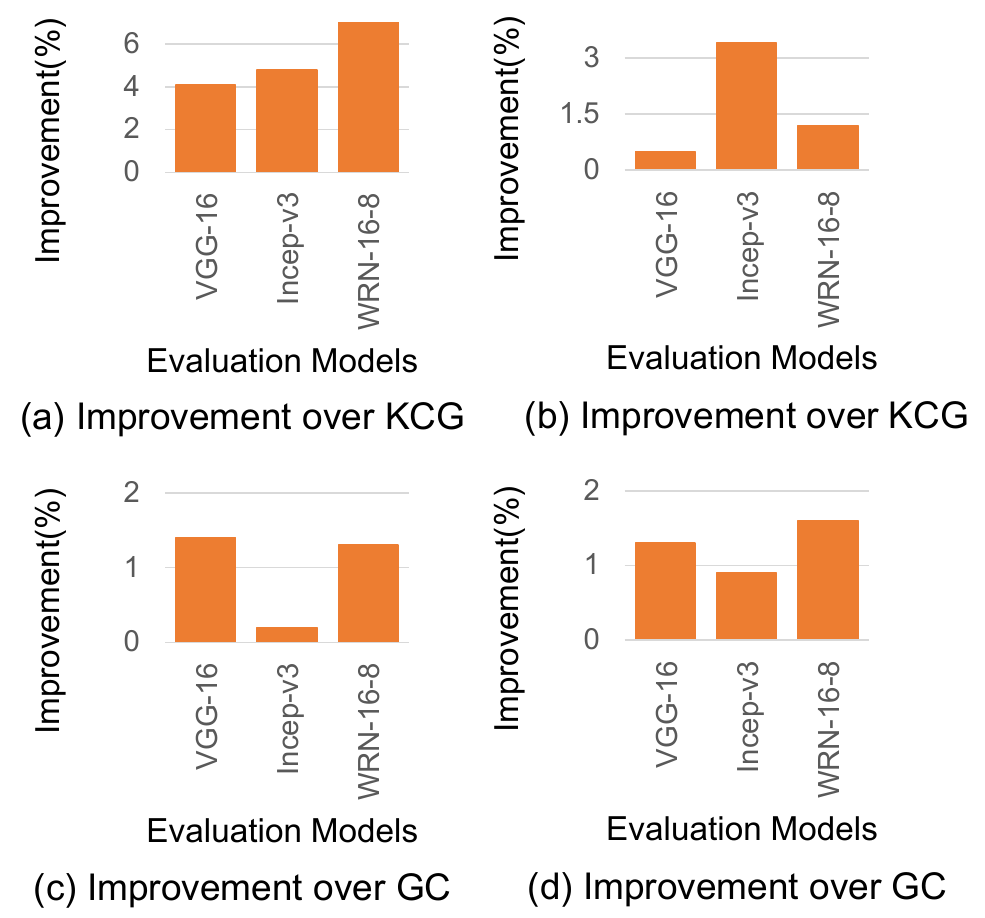}
    \caption{\textbf{Cross-architecture Performance improvement} of baseline+ours over baseline under the sampling rates of 1\% (a, c) and 10\% (b, d). The improved versions consistently outperform baselines, suggesting that increasing the diversity of CDS in the coreset facilitates generalizability across model architectures.
}
    \label{fig:cross}
\end{figure}

\subsection{Results of Cross-architecture Generalization}
We conduct cross-architecture experiments to examine whether our method benefits the model generalization. We do experiments on two methods (K-Center Greedy(KCG), Graph Cut(GC)) with three representative architectures (VGG-16~\cite{simonyan2014very}, Inception-v3(Incep-v3)~\cite{szegedy2016rethinking} and WideResNet-16-8(WRN-16-8)~\cite{zagoruyko2016wide}) under two sampling rates (1\% and 10\%). All other unspecified settings are the same as in the same-architecture experiments (\ie obtaining coresets using ResNet-18 and evaluating coresets with ResNet-18). In Figure~\ref{fig:cross}, the horizontal axis represents the model on which coresets are evaluated. All cross-architecture experiments used the ResNet-18 to obtain coresets. We can see that our method improves baselines stably, regardless of which model architecture is used to perform the evaluation. Thus, enriching the CDS diversity of the coresets facilitates cross-architecture generalizability.

\begin{figure}[!t]
    \centering
    \includegraphics[width=0.8\columnwidth]{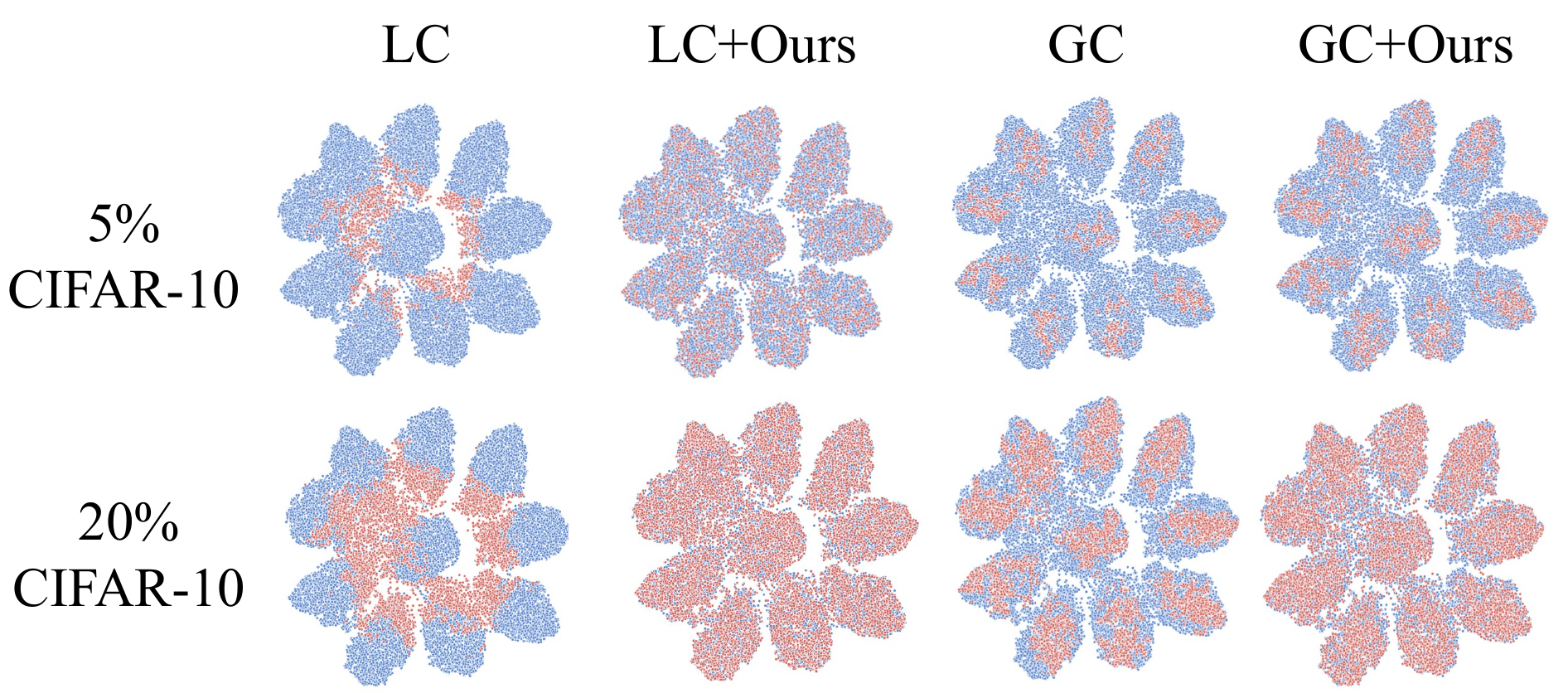}
    \caption{\textbf{tSNE embeddings.} \textcolor{red}{Selected data} are shown in red, \textcolor{blue}{unselected data} are shown in blue.
}
    \label{fig:tsnes}
    \vspace{-5mm}
\end{figure}

\subsection{TSNE embeddings of coresets}
\label{sec:tsne}
Figure~\ref{fig:tsnes} illustrates tSNE embeddings of selected and unselected data. It shows that existing methods do not handle subset diversity well, while our method motivates them to capture diversity adequately, thus boosting model performance remarkably well.

\end{document}